\documentclass[10pt,twocolumn,letterpaper]{article}

\usepackage[pagenumbers]{cvpr} 

\usepackage{booktabs}   
\usepackage{multirow}   
\usepackage{colortbl}   
\usepackage{xcolor}     
\definecolor{lightblue}{RGB}{230,245,255}  

\definecolor{cvprblue}{rgb}{0.21,0.49,0.74}
\usepackage[pagebackref,breaklinks,colorlinks,allcolors=cvprblue]{hyperref}

\title{\textit{LiteVGGT}: Boosting Vanilla VGGT via Geometry-aware Cached Token Merging
}

\author{
Zhijian Shu\textsuperscript{1,2,3}\footnotemark[1] \quad
Cheng Lin\textsuperscript{5} \quad
Tao Xie\textsuperscript{2,4} \quad
Wei Yin\textsuperscript{2} \quad
Ben Li\textsuperscript{7} \quad 
Zhiyuan Pu\textsuperscript{7}\\ 
Weize Li\textsuperscript{6} \quad 
Yao Yao\textsuperscript{3} \quad
Xun Cao\textsuperscript{3} \quad
Xiaoyang Guo\textsuperscript{2} \quad
Xiao-Xiao Long\textsuperscript{3}\footnotemark[2]\\[2mm]
\textsuperscript{1}Nanjing University of Posts and Telecommunications\quad 
\textsuperscript{2}Horizon Robotics\\
\textsuperscript{3}Nanjing University \quad
\textsuperscript{4}Zhejiang University \quad
\textsuperscript{5}Macau University of Science and Technology \\
\textsuperscript{6}TARS Robotics \quad
\textsuperscript{7}China Mobile Zijin Innovation Institute
}

\begin{document}

\twocolumn[{
	\renewcommand\twocolumn[1][]{#1}
        \vspace{-10mm}
	\maketitle
	\vspace{-11mm}
	\begin{center}
        \includegraphics[width=\textwidth]{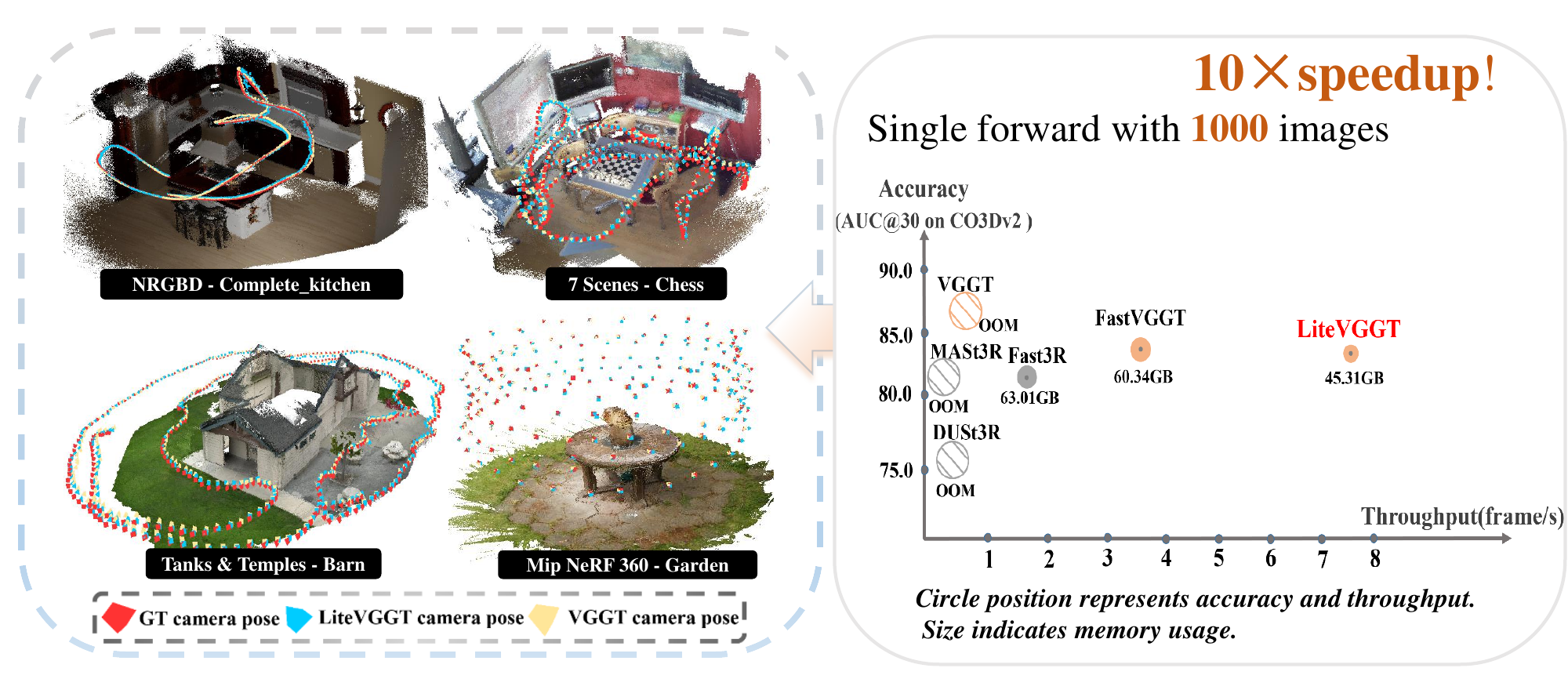}
	\end{center}
	\vspace{-10mm}
	\captionof{figure}{For 1000 input images, LiteVGGT achieves a 10× speedup over VGGT while maintaining high accuracy in camera pose and point cloud prediction. Its scalability and robustness make large-scale scene reconstruction more efficient and reliable.
}
	\label{fig:teaser}
	\vspace{3mm}
}]

\footnotetext[1]{Intern at Horizon Robotics and Nanjing University.}
\footnotetext[2]{Corresponding author.}

\begin{abstract}

3D vision foundation models like Visual Geometry Grounded Transformer (VGGT) have advanced greatly in geometric perception. However it is time-consuming and memory-intensive for long sequences, limiting application to large-scale scenes beyond hundreds of images. To address this, we propose LiteVGGT, achieving up to 10× speedup and substantial memory reduction, enabling efficient processing of 1000-image scenes. We derive two key insights for 3D reconstruction: 1) tokens from local image regions have inherent geometric correlations, leading to high similarity and computational redundancy; 2) token similarity acroses adjacent network layers remains stable, allowing for reusable merge decisions. Guided by these, we design a simple yet efficient strategy, dubbed geometry-aware cached token merging . We analyze each token’s geometric importance, optimizing anchor token selection to better preserve key information for reconstruction. We also cache and reuse merge indices across layers, substantially reducing latency with minimal accuracy impact. This strategy retains VGGT’s core performance, enabling efficient fine-tuning and FP8 quantization for further gains. Extensive experiments validate LiteVGGT’s effectiveness, scalability, and robustness. Project page is available at \url{https://garlicba.github.io/LiteVGGT/}.

\end{abstract}    

\section{Introduction}
\label{sec:intro}

Multi-view 3D reconstruction is critical for key real-world applications like autonomous navigation, augmented reality, and robotics\cite{mur2015orb, thrun2002probabilistic}, where efficiency and end-to-end capability matter most. Recent feedforward models have revolutionized the field with single-pass 3D structure prediction—avoiding the complexity of traditional Multi-View Stereo (MVS~\cite{cao2022mvsformer, wang2022efficient, cheng2024adaptive,  yao2018mvsnet, gu2020cascade, xu2022multi, cheng2022exploiting}) and computational cost of Neural Radiance Fields (NeRF)\cite{mildenhall2021nerf, takikawa2022variable, Zhang20arxiv_nerf++, cao2022mobiler2l, Park20arxiv_nerfies, bhalgat23contrastive}. Among them, Visual Geometry Grounded Transformer (VGGT)\cite{wang2025vggt} is a landmark foundation model: it handles arbitrary-length image sequences, directly predicts core 3D attributes (camera parameters, depth, point clouds, tracking features) without explicit geometric priors, and achieves SOTA performance across 3D tasks—even outperforming specialized methods. 

Despite its strengths, VGGT’s transformer architecture faces critical efficiency bottlenecks for long sequences. Its frame-global attention concatenates all tokens for full-sequence self-attention, leading to quadratic computation and memory complexity. For large-scale scenes, this inefficiency is prohibitive: Vanilla VGGT often OOMs with just 500 images, and even optimized VGGT* (with redundant memory-heavy operations removed) still takes 20 minutes on an NVIDIA H20 GPU. This prohibitive scaling makes VGGT impractical for large-scale 3D reconstruction, demanding targeted optimizations that preserve its single-pass advantage.


To address VGGT’s long-sequence inefficiency, concurrent works follow three optimized directions with notable limitations: sequential input handling (e.g., StreamVGGT~\cite{zhuo2025streaming}) sacrifices single-pass end-to-end capability; model quantization (e.g., QuantVGGT\cite{optquantized}) requires time-consuming cross-scene calibration, harming generality; token merging (e.g., FastVGGT~\cite{shen2025fastvggt}) adopts generic strategies designed for semantic/generic visual tokens (in LLMs/VLMs/diffusion models), ignoring that VGGT’s tokens are tightly geometrically coupled—carrying 3D information with one-to-one correspondence to image patches and point clouds. This oversight leads to key geometric detail loss and residual redundancy.

In contrast, VGGT’s tokens are tightly geometrically coupled: they carry 3D geometric information from local image regions, with one-to-one mapping to image patches and 3D point clouds.  Generic token merging ignores this uniqueness, treating them as semantic tokens from other domains (e.g., LLMs, diffusion models), resulting in lost geometric details, subpar reconstruction quality, and residual redundancy.

To tackle these limitations, we first conduct in-depth analysis of VGGT’s bottlenecks and token characteristics. We identify that the core bottleneck lies in the global block, where concatenating all tokens for self-attention leads to excessive computation and memory consumption. Further observations reveal two key 3D-specific insights: 1) Tokens from local image regions exhibit inherent geometric correlations, resulting in high cross-frame similarity and significant computational redundancy; 2) Token similarity remains stable across adjacent network layers, enabling reusable merge decisions. 

Guided by these insights, we propose a geometry-aware cached token merging strategy, tailored to 3D reconstruction, to preserve key geometric information while reducing redundancy. Specifically, we first construct a geometry-aware feature map by fusing pixel gradient (capturing edges/textures) and token variance (measuring semantic-geometric variability) to quantify each token’s geometric importance. We then partition tokens into three categories: GA Tokens (top 10\% high-importance tokens, retained to preserve critical geometric details), dst Tokens (spatially balanced anchor tokens), and src Tokens (redundant tokens to be merged). Src tokens are merged into the most similar dst tokens via cosine similarity-based feature averaging, with token replication (unmerging) restoring sequence length for dense prediction. To eliminate redundant computation, we cache and reuse merge indices across adjacent layers. Additionally, this strategy retains VGGT’s core performance, laying the foundation for efficient fine-tuning and FP8 quantization to further enhance efficiency.


To summarize, we make the following contributions:
\begin{itemize}
\item Propose LiteVGGT with a geometry-aware cached token merging strategy for efficient redundancy reduction while preserving reconstruction quality.
\item Complemented by fine-tuning and FP8 quantization, it delivers up to 10× speedup and substantial memory savings over VGGT.
\item Extensive experiments validate its efficiency, generality, and scalability—matching VGGT’s performance while enabling large-scale scene reconstruction.
\end{itemize}

\section{Related work}

\begin{figure*}[t]
  \centering
  \includegraphics[width=1\linewidth]{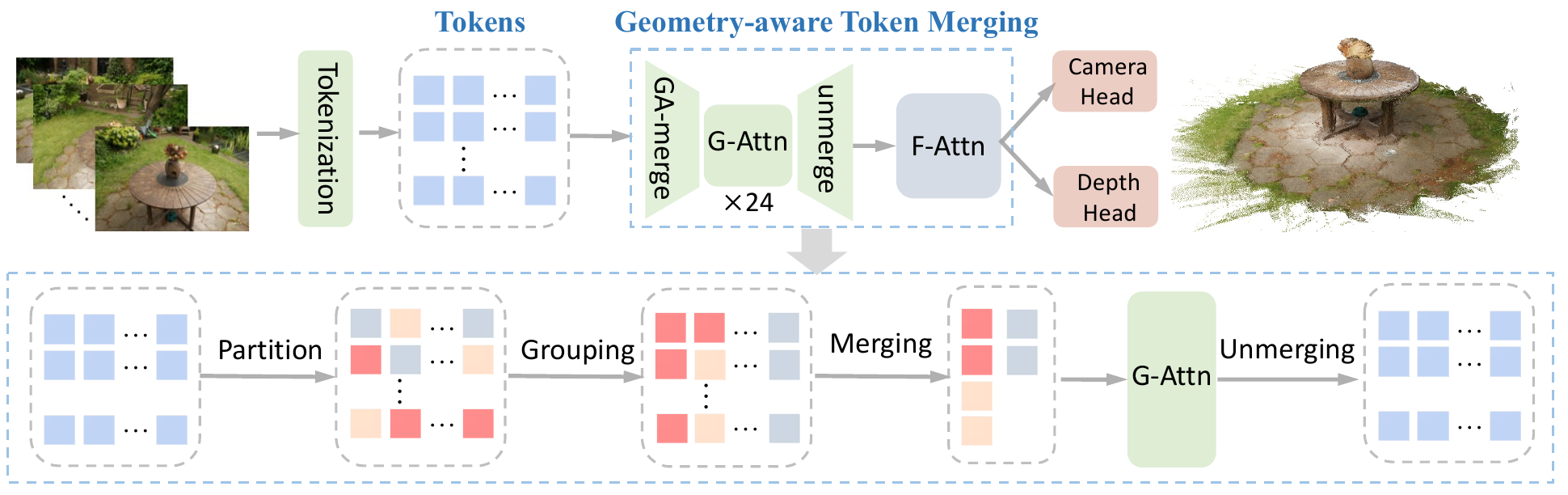}
  \caption{\textbf{Architecture Overview.}
We augment VGGT by placing a Geometry-aware Token Merging module on both sides of its global attention. 
Within GA-merge, tokens are partitioned by the GA map, grouped and merged to reduce redundancy, and after global attention the merged tokens are unmerged back to the original layout and passed to the subsequent frame-attention layers.
 }
  \label{fig:pipeline}
\end{figure*}

\subsection{Feedforward 3D Foundation Models}
Building on conventional 3D reconstruction, recent end-to-end learning-based methods~\cite{yang2025fast3r,reizenstein2021common,zhang2024gs,zhang2024monst3r,szymanowicz2025flash3d,smart2024splatt3r,yang2025fast3r,long2024wonder3d,murai2025mast3r,lu2025align3r} encode scene priors via neural networks, significantly boosting robustness and cross-dataset generalization. DUST3R~\cite{wang2024dust3r} pioneered regressing view-consistent 3D point maps from two RGB images without camera calibration, while its successor MASt3R~\cite{murai2025mast3r} refines metric scale estimation via confidence-weighted losses.
For single-pass dense-view interaction, Fast3R~\cite{yang2025fast3r} extends to N-view processing with Flash-Attention and parallel fusion, supporting 1000+ image inference and SOTA accuracy in seconds.
State-of-the-art VGGT~\cite{wang2025vggt} (1.2B parameters) jointly predicts camera parameters, depth, and point maps in one forward pass, outperforming task-specific methods. However, its quadratic complexity, offline inference, and large size hinder real-world deployment. Its transformer architecture enables compression—our core focus.

\subsection{Token Merging}

Visual token merging originated in Vision Transformers (ViTs~\cite{dosovitskiy2020image}) as a training-free technique to accelerate inference by merging redundant tokens, later extended to diffusion models~\cite{rombach2021highresolution}, video understanding~\cite{zhao2023learning}, and vision-language models (VLMs~\cite{zhang2023video, song2024moviechat, lin2025vlog}) for computational cost reduction. It partitions tokens into source (src) and destination (dst) via region-based random sampling, merging similar pairs or averaging via token pooling~\cite{nawrot2023efficient,pietruszka2022sparsifying}. Key works include TokenLearner~\cite{ryoo2021tokenlearner} (MLP-based token selection), ToMe~\cite{bolya2022token} (per-block token reduction), PuMer~\cite{cao2023pumer} (VLM-oriented token pruning/merging), and ToMeSD~\cite{bolya2023token} (Stable Diffusion adaptation with unmerge for dense prediction). Despite effectiveness, these methods ignore long-sequence feed-forward 3D reconstruction’s demand for spatial precision and temporal coherence—stemming from their random sampling. This motivates our LiteVGGT’s geometry-aware token merging, tailored for 3D reconstruction.

\begin{figure}[t]
  \centering
  \includegraphics[width=1\linewidth]{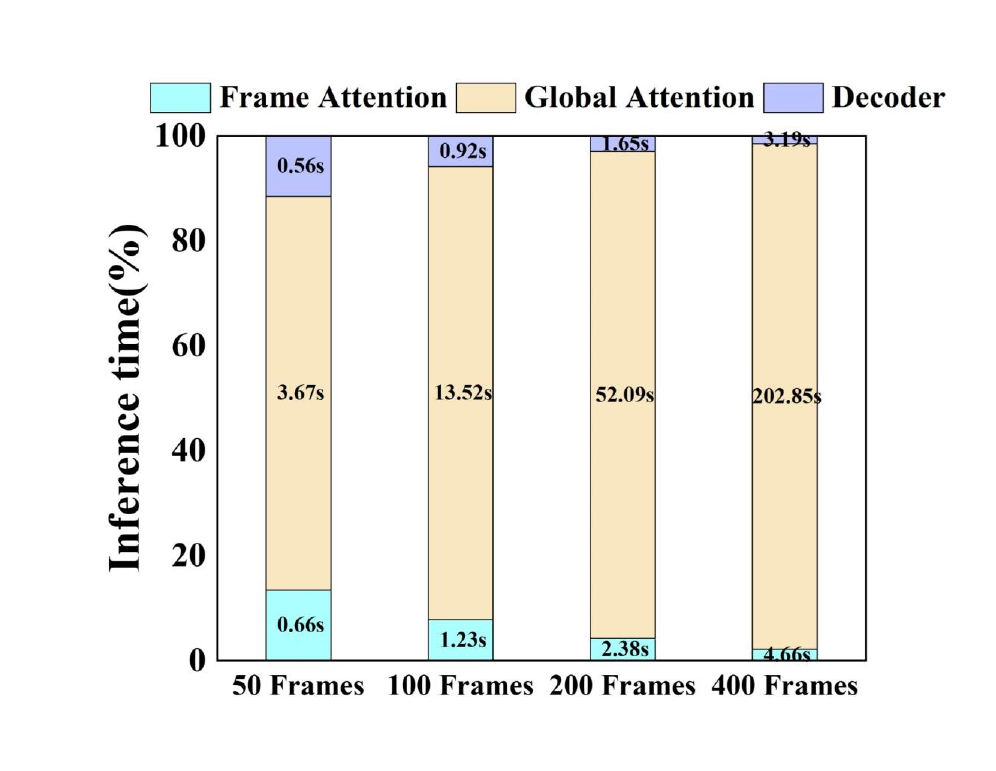}
  \vspace{-8mm}
  \caption{ Latency analysis of the VGGT components. As the number of images increases, Global Attention gradually dominates the inference time.}
  \label{fig:global_time}
\end{figure}

\begin{figure}[t]
  \centering
  \includegraphics[width=1\linewidth]{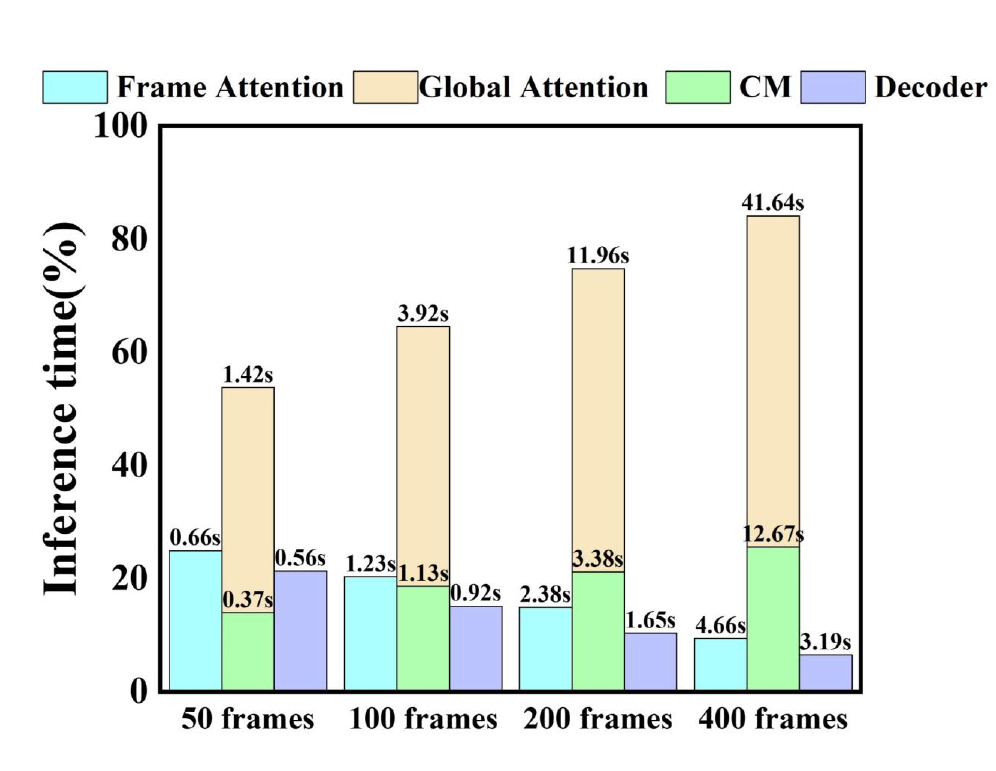}
  \vspace{-8mm}
  \caption{Latency breakdown after introducing token merging. CM denotes merge index computation latency, which becomes a bottleneck for long sequences—addressed by our caching strategy.}
  \label{fig:latency}
\end{figure}

\section{Method}
We introduce LiteVGGT, a lightweight version of VGGT that achieves a 10× speedup while maintaining high accuracy. We begin by analyze architecture and bottleneck of VGGT in Sec.~\ref{subsec:bottle}, followed by the common-used naive token merging strategy in Sec.~\ref{subsec: naive}. In Sec.~\ref{subsec: cues}, we analyze the geometric cues learned by 3D foundation models, and in Sec.~\ref{subsec: ga_merge}, we describe our geometry-aware cached token merge strategy. Finally, in Sec.~\ref{subsec: finetune}, we present the details of fine-tuning and FP8 quantization for further improvement.

\subsection{VGGT Architecture and Bottleneck Analysis}
\label{subsec:bottle}

\textbf{Core Architecture.} Visual Geometry Grounded Transformer (VGGT) ~\cite{wang2025vggt} is a 3D foundation model for arbitrary-length image sequences, predicting key scene attributes. It tokenizes RGB frames via a pretrained backbone (e.g., DINOv2~\cite{oquab2023dinov2}) to generate patch tokens, augmented with 5 special tokens (1 camera + 4 register) per frame for aggregating 3D attributes (e.g., camera parameters, geometry). All tokens are fed into a 24-layer frame-global attention mechanism, with outputs directed to task-specific prediction heads.



\textbf{Bottleneck Analysis.} VGGT’s speed bottleneck stems from its frame-global attention: to ensure inter-frame consistency, it concatenates tokens across all images for full-sequence self-attention, leading to quadratic complexity with token count—rendering it impractical for large-scale scenes as inputs scale to hundreds/thousands of frames (Figure \ref{fig:global_time}. This issue is exacerbated by computational redundancy: its vision backbone extracts local semantic-geometric features, resulting in highly similar tokens across overlapping regions of scene frames, and attention patterns remain stable across adjacent layers (Figure \ref{fig:sim}), indicating consistent inter-layer token similarity. These observations confirm significant redundancy in VGGT’s Global Attention, motivating our 3D reconstruction-tailored token merging strategy.

\begin{figure}[t]
  \centering
  \includegraphics[width=\linewidth]{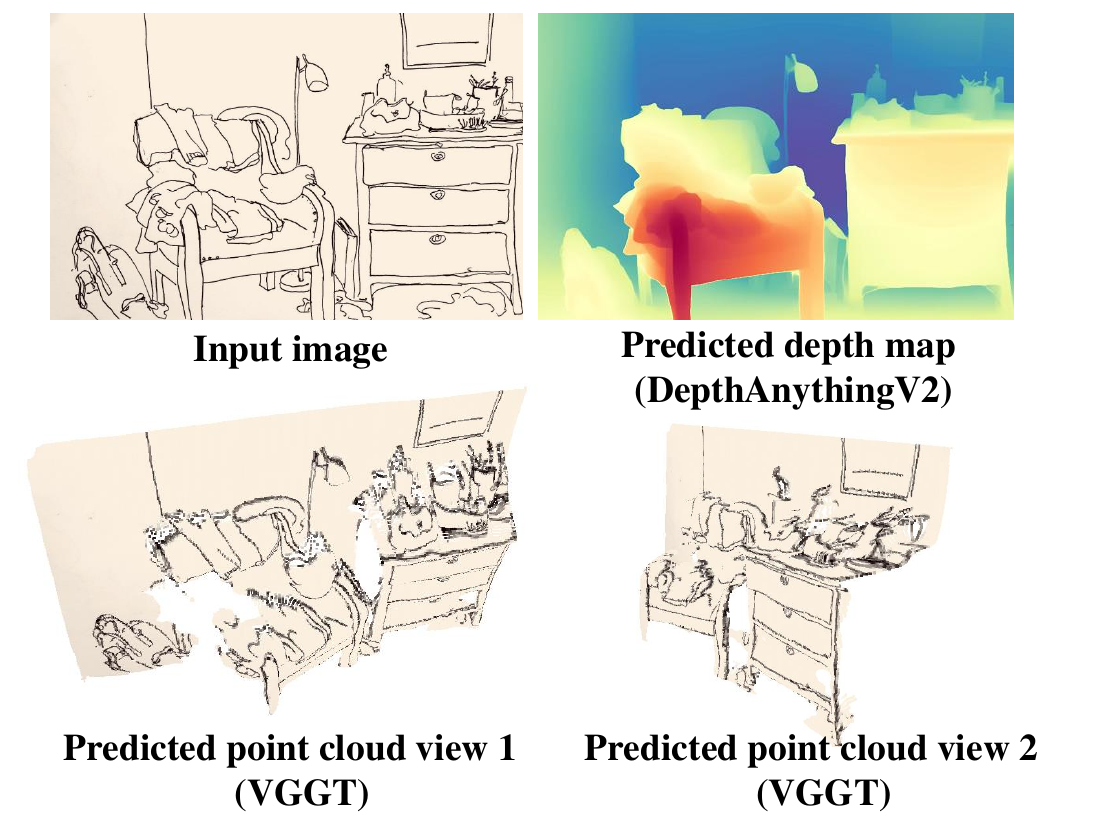}
  \caption{Experiment about geometric cues. VGGT and DepthAnythingV2~\cite{yang2024depth} still produce reasonable geometric results of input edge map.}
  \label{fig:3d_reason}
\end{figure}

\subsection{Naive Token Merging}
\label{subsec: naive}

Token merging is a widely adopted approach to reduce transformer sequence length, but generic strategies are ill-suited for VGGT and 3D reconstruction due to their ignorance of geometric constraints. For LLMs or diffusion models, such methods typically partition tokens into destination (dst) and source (src) sets via random sampling or fixed-stride pooling (e.g., ToMe \cite{bolya2022token}), prioritizing computational reduction over task-specific demands, tokens in these domains encode semantic or generic visual information with no inherent geometric binding.

For VGGT, these generic strategies fail to account for 3D reconstruction’s unique requirement: its tokens are tightly coupled with image regions and 3D point clouds. Random or fixed-stride partitioning risks merging high-information geometric tokens (e.g., edges, textures) with low-information ones, leading to critical detail loss. Additionally, recomputing merge indices for all network layers introduces redundant overhead. These limitations motivate our geometry-aware and cached token merging strategy, tailored to preserve 3D geometric information while maximizing efficiency.




\subsection{Geometric Cues for 3D Reasoning}
\label{subsec: cues}

Traditional 3D reconstruction relies heavily on texture patterns for feature matching, but modern learning-based models like VGGT and DepthAnything-V2 \cite{yang2024depth} exhibit distinct geometric perception priors. As shown in Fig.~\ref{fig:3d_reason}, we fed pure edge maps (stripped of all texture and photometric information) into both models—surprisingly, they still generated coherent, geometrically plausible reconstructions. This confirms that 3D vision models depend heavily on structural contour (edge) information for geometric inference, even without photorealistic details. 

This key insight that edge-rich regions (high gradients, high semantic variance) are the backbone of 3D geometric reasoning directly motivates our Geometry-aware merging design (Section~\ref{subsec: ga_merge}). The strategy prioritizes tokens in such regions to preserve critical geometric information during merging.




\subsection{Geometry-aware Cached Token Merging}
\label{subsec: ga_merge}

To address naive token merging’s limitations, we propose \textit{Geometry-aware (GA) Cached Token Merging} tailored to 3D reconstruction’s geometric preservation demand with three core innovations: geometry-aware token prioritization, partitioning, and cached merging/unmerging (Fig. \ref{fig:pipeline}).

\begin{figure}[t]
  \centering
  \includegraphics[width=1\linewidth]{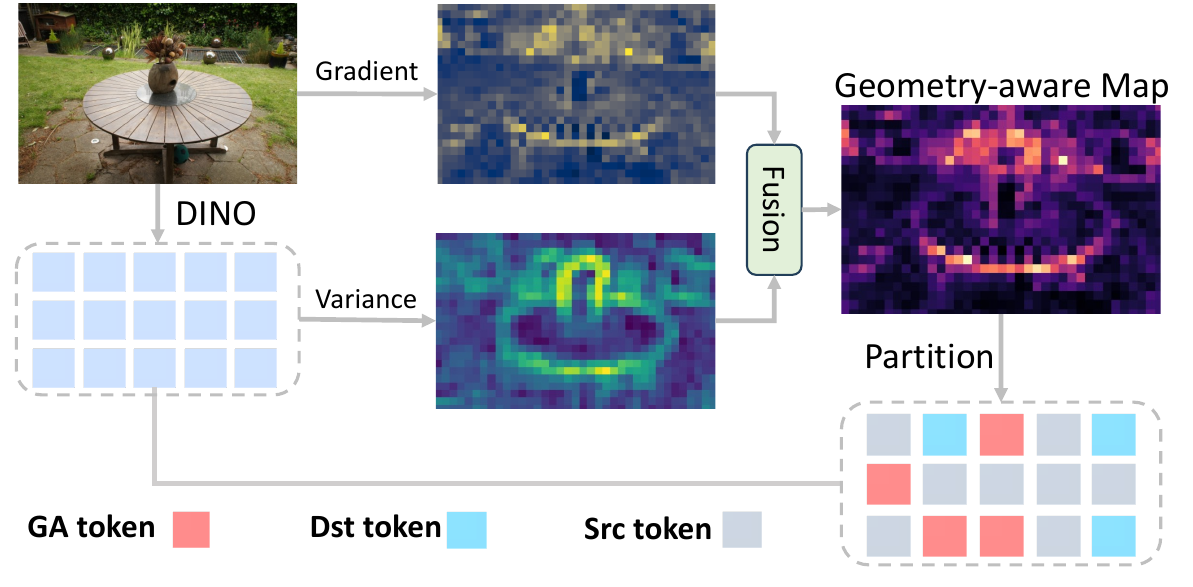}
  \caption{The illustration of Geometry-aware Cached Token Merging.}
  \label{fig:gamerge}
\end{figure}

\begin{figure*}[t]
\centering
\includegraphics[width=1\linewidth]{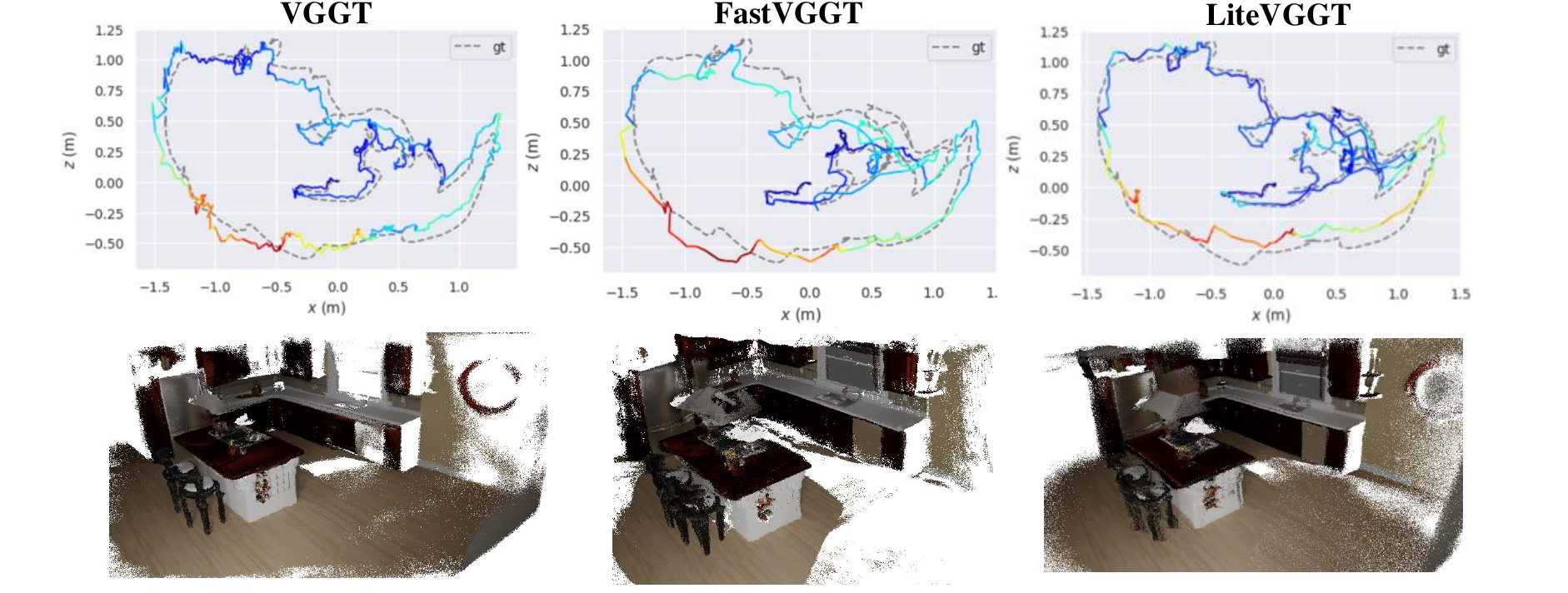}
\vspace{-5mm}
\caption{Qualitative comparisons of VGGT, FastVGGT, and LiteVGGT on camera pose estimation and point cloud reconstruction.
All point clouds are visualized under the same confidence threshold with the same number of points retained.}
\label{fig:main_result}
\end{figure*}

\begin{table*}[tp]
\centering
\setlength{\tabcolsep}{11pt} 
\renewcommand{\arraystretch}{1.2}
\resizebox{0.8\linewidth}{!}{
\begin{tabular}{lcccccccc}
\toprule
\multirow{2}{*}{\textbf{Method}} & 
\multicolumn{2}{c}{\textbf{1000 images}} & 
\multicolumn{2}{c}{\textbf{496 images}} & 
\multicolumn{2}{c}{\textbf{296 images}} & 
\multicolumn{2}{c}{\textbf{96 images}} \\
\cmidrule(lr){2-3} \cmidrule(lr){4-5} \cmidrule(lr){6-7} \cmidrule(lr){8-9}
 & CD $\downarrow$ & Time $\downarrow$ 
 & CD $\downarrow$ & Time $\downarrow$
 & CD $\downarrow$ & Time $\downarrow$
 & CD $\downarrow$ & Time $\downarrow$ \\
\midrule
Fast3R \cite{yang2025fast3r} & 0.692 & 563.7s & 0.689 & 143.7s & 0.711 & 41.3s & 0.720 & 5.9s \\
CUT3R \cite{wang2025continuous} & 0.780 & 112.2s & 0.774 & 26.7s & 0.754 & 13.5s & 0.791 & 5.1s \\
VGGT \cite{wang2025vggt} & OOM & OOM & OOM & OOM & 0.417 & 121.2s & 0.420 & 16.6s \\
VGGT$^*$ \cite{wang2025vggt} & 0.485 & 1275.1s & \underline{0.424} & 329.7s & \underline{0.415} & 120.9s & 0.418 & 16.7s \\
FastVGGT \cite{shen2025fastvggt} & \underline{0.436} & 258.3s & \underline{0.424} & 78.4s & 0.423 & 32.8s & \underline{0.409} & 6.4s \\
LiteVGGT & \textbf{0.428} & 127.2s & \textbf{0.392} & 37.9s & \textbf{0.365} & 16.6s & \textbf{0.329} & 3.5s \\
\bottomrule
\end{tabular}
}
\caption{Quantitative results of \textbf{point cloud reconstruction on the ScanNet-50 dataset}\cite{dai2017scannet}. OOM denotes out-of-memory. Bold indicates the best result, and underline denotes the second best.}
\label{tab:recon_scannet50}
\end{table*}

\textbf{Geometry-aware Feature Map.} To quantify the geometric importance of each token, we design a geometry-aware feature map that fuses two lightweight yet effective cues:
\begin{itemize}
    \item \textit{Pixel Gradient (Grad Map)}: Captures edges/texture boundaries via the Sobel operator (horizontal/vertical intensity changes), downsampled to token granularity to highlight edge-rich regions critical for 3D reasoning.
    \item \textit{Token Variance (Var Map)}: Measures semantic variability by rearranging tokens into 2D grids, computing local average-pooled variance to distinguish uniqueness (e.g., textured surfaces) from smooth regions (e.g., blank walls).
\end{itemize}

\begin{figure}[tp]
\centering
\includegraphics[width=1\linewidth]{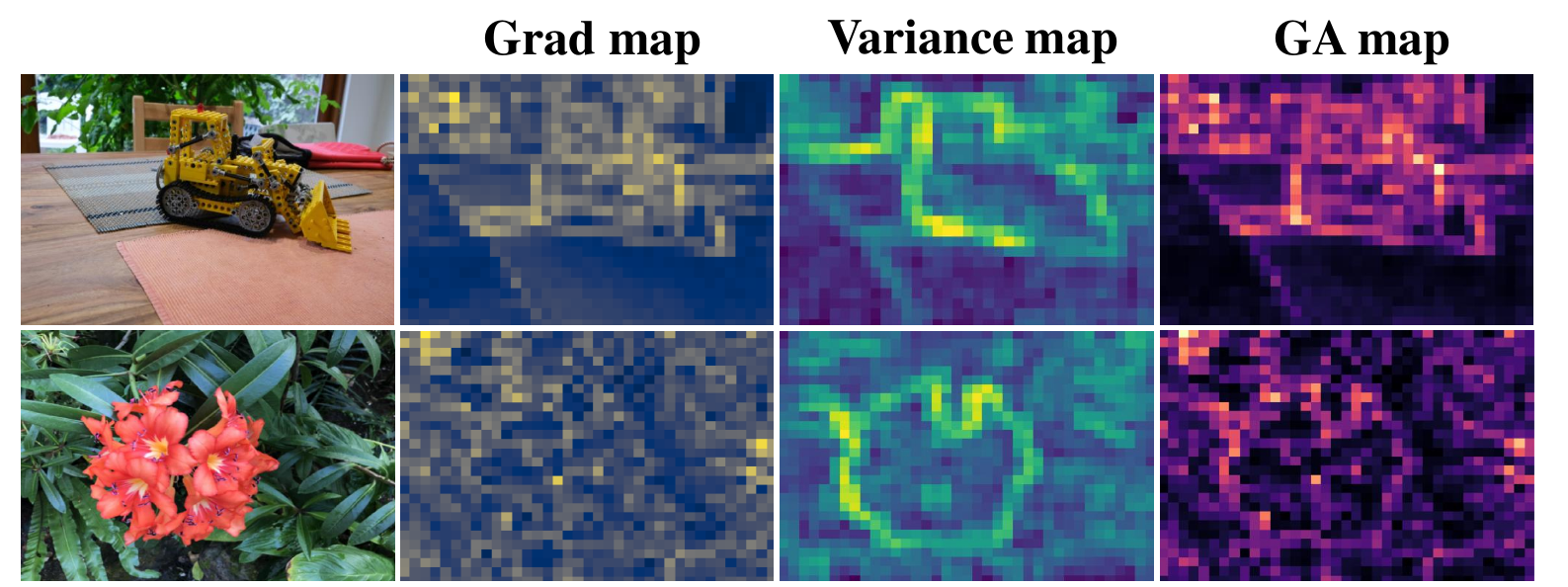}
\caption{Visualization of pixel gradients (Grad map), token variance (Variance map), and the fused Geometry-aware map (GA map). }
\label{fig:info_map}
\end{figure}

The two types of maps are fused together to form the final Geometry-aware map, integrating both geometric structure and semantic variability:
\[
\Psi_{\text{GA}} = \alpha \cdot \text{norm}(\Psi_g) + \beta \cdot \text{norm}(\Psi_v)
\]
where \(\Psi_g\) (Grad Map) and \(\Psi_v\) (Var Map) are normalized, \(\alpha/\beta\) balance their contributions. As visualized in Fig. \ref{fig:info_map}, \(\Psi_{\text{GA}}\) clearly distinguishes high-information tokens (edges/textures) from low-redundancy ones (smooth surfaces), laying the foundation for targeted partitioning.


\begin{figure*}[tp]
  \centering
  \includegraphics[width=1.0\linewidth, height=6cm]{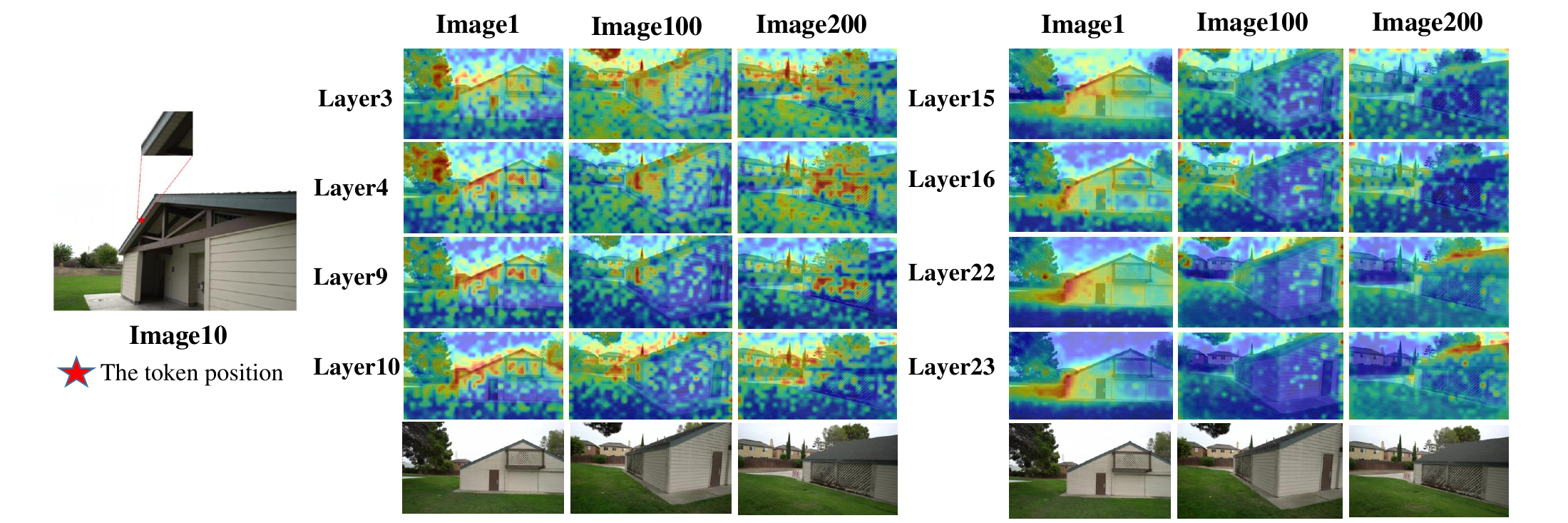}
  \caption{The figure shows the attention between a randomly selected token from Image10 (as the query) and all other tokens (as values) from three images. 
  Each row corresponds to different layers, and each column shows attention patterns across images, demonstrating high similarity between tokens across frames. The similarity of attention maps across adjacent layers indicates the stability of token similarity between layers.}
  \label{fig:sim}
\end{figure*}

\textbf{GA Token Partitioning.} Guided by the GA map, we partition tokens into three categories to balance efficiency (token reduction) and reconstruction quality (geometric detail preservation), as shown in Fig. \ref{fig:gamerge}:
\begin{itemize}
    \item \textit{GA Tokens}: The top 10\% highest-scoring tokens per frame in the GA map. These tokens correspond to critical geometric/semantic details (e.g., object edges, textured regions) and are \textit{excluded from merging} to avoid core information loss.
    \item \textit{Dst Tokens}: Merge anchors selected for spatial coherence and efficiency: (1) All first-frame tokens (VGGT’s world coordinate anchor for inter-frame consistency); (2) One token per 2×2 grid in other frames (lowest GA score, targeting smooth, low-information regions to maximize merge efficiency).
  \item \textit{Src Tokens}: All remaining tokens (not selected as GA Tokens or dst Tokens), which are designated for merging into the most similar dst tokens.
\end{itemize}

\textbf{Cached Token Merging}  
This module integrates redundant token reduction and inter-layer index caching to cut quadratic complexity while preserving geometric consistency—addressing both token redundancy and redundant computation in naive methods.  
For src tokens, we first match them to dst tokens via cosine similarity (ensuring geometric alignment). Each dst token’s feature is updated by averaging its own feature with assigned src tokens, reducing sequence length for Global Attention:  
\[
x_d' = \frac{x_d + \sum_{x_s \in \mathcal{S}_d} x_s}{1 + |\mathcal{S}_d|}
\]  
where \(x_d\) is the original dst feature, \(\mathcal{S}_d\) denotes assigned src tokens, and \(x_d'\) is the updated feature. Only \(x_d'\) is retained for subsequent layers.  

Leveraging stable inter-layer token similarity (Figure \ref{fig:sim}), we cache merge indices to avoid redundant computation: indices are computed once every 6 layers (4 total computations) and reused for intervening layers. This reduces latency by 20\% with minimal accuracy loss (Figure \ref{fig:latency}).

\textbf{Token Unmerging}  
To support VGGT’s dense outputs (e.g., depth maps, point clouds), we restore the token sequence to its original length before prediction heads. Each merged \(x_d'\) is replicated to recover all tokens in \(\{x_d\} \cup \mathcal{S}_d\), with local geometric variability regained via VGGT’s Frame Attention—ensuring no degradation in dense prediction detail.

\subsection{Fine-tuning and FP8 Quantization}
\label{subsec: finetune}
To maximize efficiency while preserving accuracy, we complement our token merging strategy with two engineering optimizations:

\textbf{Fine-tuning.} Building on VGGT’s pretrained checkpoints, we fine-tune the aggregator and prediction heads on a diverse dataset mix (Co3Dv2\cite{reizenstein2021common}, BlendMVS\cite{yao2020blendedmvs}, DL3DV\cite{ling2024dl3dv}, etc.). We sample 4–48 images per batch and train for 20K iterations on 8 H20 GPUs (about 3 days) to mitigate the accuracy loss introduced by token merging. During fine-tuning, we employ a composite learning rate schedule with a 5\% linear warm-up from $1\times10^{-6}$ to $4\times10^{-5}$, followed by a cosine decay to $7\times10^{-7}$ over the remaining 95\% of training.

\textbf{FP8 Quantization.} Using NVIDIA’s Transformer Engine, we adopt FP8 inference after weight mapping. This reduces memory footprint and latency with minimal performance degradation, enabled by our token merging strategy’s preservation of VGGT’s core feature representation.

\begin{table*}[tp]
\centering
\setlength{\tabcolsep}{6pt}
\renewcommand{\arraystretch}{1.2}
\resizebox{0.7\linewidth}{!}{
\begin{tabular}{l|ccc|c|ccc|c}
\toprule
\multirow{2}{*}{\textbf{Method}}
& \multicolumn{4}{c|}{\textbf{7 Scenes}}
& \multicolumn{4}{c}{\textbf{NRGBD}} \\
\cline{2-9}
& \textbf{Acc} $\downarrow$ & \textbf{Comp} $\downarrow$ & \textbf{NC} $\uparrow$ & \textbf{Time} $\downarrow$
& \textbf{Acc} $\downarrow$ & \textbf{Comp} $\downarrow$ & \textbf{NC} $\uparrow$ & \textbf{Time} $\downarrow$ \\
\midrule

Fast3R \cite{yang2025fast3r}
& 0.053 & 0.047 & \underline{0.613} & 58.5s
& 0.083 & 0.023 & 0.655 & 98.9s \\

CUT3R \cite{wang2025continuous}
& 0.181 & 0.088 & 0.586 & \textbf{15.8s}
& 0.304 & 0.166 & 0.581 & \textbf{27.1s}\\

VGGT \cite{wang2025vggt}
& \underline{0.021} & \textbf{0.022} & \textbf{0.614} & 133.1s
& \textbf{0.024} & \underline{0.015} & \textbf{0.730} & 233.1s \\

FastVGGT \cite{shen2025fastvggt} 
& \underline{0.021} & \textbf{0.022} & 0.604 & 36.5s
& \underline{0.029} & \textbf{0.014 }& \underline{0.704} & 70.7s \\

LiteVGGT
& \textbf{0.020} & \underline{0.024} & 0.606 & 
\underline{19.2s}
& 0.031 & 0.019 & 0.698 & \underline{36.3s} \\
\bottomrule
\end{tabular}
}
\caption{Quantitative results of \textbf{point cloud reconstruction on the 7 Scenes\cite{shotton2013scene} and NRGBD\cite{azinovic2022neural}.}}
\label{tab:recon_7scenes_nrgbd}
\end{table*}

\begin{table*}[tp]
\centering
\resizebox{0.8\linewidth}{!}{
\begin{tabular}{lccccccccc}
\toprule
\textbf{Method} & \textbf{Barn} & \textbf{Caterpillar} & \textbf{Courthouse} & 
\textbf{Ignatius} & \textbf{Meetingroom} & \textbf{Truck} & \textbf{Avg.$\uparrow$} & \textbf{Time(s)$\downarrow$} \\
\midrule
threshold & 0.050 & 0.025 & 0.015 & 0.025 & 0.050 & 0.250 & ×5 & - \\
\midrule
VGGT \cite{wang2025vggt} & 0.54 & 0.33 & OOM & 0.54 & 0.63 & 0.62 & - & - \\
VGGT$^*$ \cite{wang2025vggt} & 0.54 & 0.33 & 0.50 & 0.54 & 0.63 & 0.62 & \textbf{0.53} & 221.45s \\
FastVGGT \cite{shen2025fastvggt} & 0.37 & 0.27 & 0.44 & 0.34 & 0.43 & 0.45 & 0.38 & 66.20s \\
LiteVGGT & 0.44 & 0.26 & 0.44 & 0.30 & 0.57 & 0.41 & \underline{0.40} & 29.52s \\
\midrule
threshold & 0.100 & 0.050 & 0.030 & 0.050 & 0.100 & 0.500 & ×10 & - \\
\midrule
VGGT \cite{wang2025vggt} & 0.69 & 0.56 & OOM & 0.70 & 0.79 & 0.74 & - & - \\
VGGT$^*$ \cite{wang2025vggt} & 0.69 & 0.56 & 0.67 & 0.70 & 0.79 & 0.74 & \textbf{0.69} & 221.45s \\
FastVGGT \cite{shen2025fastvggt} & 0.56 & 0.42 & 0.61 & 0.48 & 0.61 & 0.59 & 0.54 & 66.20s \\
LiteVGGT & 0.60 & 0.43 & 0.60 & 0.45 & 0.74 & 0.59 & \underline{0.57} & 29.52s \\
\bottomrule
\end{tabular}
}
\caption{Quantitative results of \textbf{point cloud reconstruction on the Tanks \& Temples\cite{knapitsch2017tanks}}. We report the F1-score and average inference time. ×5 refers to expanding the official threshold values by a factor of 5.}
\label{tab:recon_tnt}
\end{table*}

\section{Experiments}

\subsection{Implementation Details}
\textbf{Datasets.}
We evaluate two core tasks across diverse datasets: 3D reconstruction is tested on ScanNet-50 \cite{dai2017scannet} (50 scenes, following FastVGGT protocol with variable sequence lengths), 7Scenes \cite{shotton2013scene} and NRGBD \cite{azinovic2022neural} (indoor sequences with keyframes sampled every 3 frames), Tanks \& Temples \cite{knapitsch2017tanks} (6 training scenes for large-scale outdoor testing), DTU \cite{jensen2014large} (48 uniformly sampled frames per object), and CO3Dv2 \cite{reizenstein2021common} (8 uniformly sampled images per scene); camera pose estimation is evaluated on CO3Dv2 \cite{reizenstein2021common}, Tanks \& Temples \cite{knapitsch2017tanks}, and DTU \cite{jensen2014large} with the same sampling strategy as the reconstruction task.

\textbf{Evaluation Metrics.}
In experiments, we use the Chamfer Distance (CD), 
Accuracy (Acc), 
Completeness (Comp), 
and Normal Consistency (NC) 
to evaluate the point cloud quality. For large-scale outdoor scenes, we additionally adopt the F1 score to quantify the percentage of matched points between predicted and ground-truth point clouds, with F1 thresholds scaled ×5/×10 to filter minor noise—accounting for VGGT’s speed-accuracy trade-off compared to optimization-based methods.
In camera pose evaluations, we report the AUC of
the relative pose error within (5°, 15°, 30°), where the pose error is the maximum between the rotation and translation angular error.

\begin{table}[h]
\centering
\setlength{\tabcolsep}{5pt}
\renewcommand{\arraystretch}{1.15}
\resizebox{1.0\linewidth}{!}{
\begin{tabular}{lccc|ccc}
\toprule
\multirow{2}{*}{\textbf{Method}} 
& \multicolumn{3}{c|}{\textbf{DTU Reconstruction}} 
& \multicolumn{3}{c}{\textbf{CO3Dv2 Pose Estimation} } \\
\cline{2-7}
& \textbf{Acc. $\downarrow$} & \textbf{Comp. $\downarrow$} & \textbf{Over. $\downarrow$} 
& \textbf{AUC@30$\uparrow$} & \textbf{AUC@20$\uparrow$} & \textbf{AUC@15$\uparrow$} \\
\midrule

Fast3R \cite{yang2025fast3r} 
& 3.712 & 1.412 & 2.562 
& 81.3 & 74.2 & 69.7 \\

CUT3R \cite{wang2025continuous}
& 1.428 & 1.396 & 1.412
& 82.5 & 77.3 & \underline{73.8} \\

VGGT \cite{wang2025vggt}
& \textbf{0.508} & \textbf{0.561} & \textbf{0.534}
& \textbf{86.3} & \textbf{81.4} & \textbf{76.8} \\

FastVGGT \cite{shen2025fastvggt}
& 0.824 & \underline{0.655} & 0.739
& \underline{83.4} & \underline{77.6} & 71.9 \\

LiteVGGT
& \underline{0.652} & 0.780 & \underline{0.716}
& 83.2 & 77.3 & 71.9 \\

\bottomrule
\end{tabular}
}
\caption{Quantitative results of \textbf{point cloud reconstruction on the DTU dataset\cite{jensen2014large} }(left) and \textbf{camera pose estimation on CO3Dv2\cite{reizenstein2021common} }(right).}
\label{tab:combined_dtu_co3d}
\end{table}

\subsection{3D Reconstruction}

\textbf{Quantitative comparisons.} 
On the ScanNet-50 dataset (Table \ref{tab:recon_scannet50}), CUT3R achieves the fastest runtime but at the cost of notable accuracy degradation. In contrast, LiteVGGT delivers the lowest Chamber Distance (CD) error and a 10× speed-up over VGGT with 1000 input images. The results on 7Scenes and NRGBD (Table \ref{tab:recon_7scenes_nrgbd}) further show that LiteVGGT offers accuracy comparable to the best methods while achieving substantially higher efficiency, reflecting a strong accuracy–efficiency trade-off. Moving to object-level reconstruction, Table \ref{tab:combined_dtu_co3d} (left) demonstrates that LiteVGGT maintains competitive performance on DTU, performing slightly below VGGT but clearly surpassing FastVGGT. Finally, on large-scale outdoor scenes from Tanks \& Temples (Table \ref{tab:recon_tnt}), LiteVGGT outperforms FastVGGT under both threshold settings and achieves nearly a 10× speed-up over VGGT. Although its F1 score is slightly lower than VGGT, the combined accuracy and efficiency results clearly indicate strong scalability and generalization for large-scene reconstruction.

\begin{table*}[tp]
\centering
\resizebox{0.9\linewidth}{!}{
\begin{tabular}{lcccccccc}
\toprule
\textbf{Method} & \textbf{Barn} & \textbf{Caterpillar} & \textbf{Courthouse} & 
\textbf{Ignatius} & \textbf{Meetingroom} & \textbf{Truck} & \textbf{Avg.$\uparrow$} & \textbf{Time (s)$\downarrow$} \\
\midrule
VGGT \cite{wang2025vggt}& 91.3 & 89.0 & OOM & 88.9 & 90.0 & 94.1 & - & - \\
VGGT$^*$ \cite{wang2025vggt}& 91.3 & 89.0 & 81.8 & 88.8 & 90.0 & 94.1 & \textbf{89.2} & 221.45s \\
FastVGGT \cite{shen2025fastvggt}  & 90.1 & 86.3 & 81.8 & 88.0 & 87.6 & 93.5 & 87.9 & 66.20s \\
LiteVGGT & 88.5 & 89.3 & 81.7 & 88.1 & 87.8 & 92.6 & \underline{88.0} & \textbf{29.52s} \\
\bottomrule
\end{tabular}
}
\caption{Quantitative results of \textbf{camera pose estimation on Tanks \& Temples\cite{knapitsch2017tanks}}, evaluated using AUC@30.}
\label{tab:pos_tnt}
\end{table*}

\textbf{Qualitative comparisons.}
As shown in the bottom part of Figure \ref{fig:main_result}, we visualize the reconstructed point clouds of VGGT, FastVGGT, and LiteVGGT on the complete\_kitchen scene from NRGBD. Although LiteVGGT produces slightly less detailed geometry compared to VGGT, it preserves stronger overall completeness and geometric consistency. These qualitative results highlight that LiteVGGT maintains high reconstruction quality while achieving a substantial speed-up.

\begin{table}[tp]
\centering
\renewcommand{\arraystretch}{1.0}
\setlength{\tabcolsep}{3pt}
\resizebox{0.85\linewidth}{!}{
\begin{tabular}{lccc}
\toprule
\textbf{Method} & \textbf{AUC@30} $\uparrow$ & \textbf{AUC@15} $\uparrow$ & \textbf{AUC@5} $\uparrow$ \\
\midrule
VGGT \cite{wang2025vggt} & 94.3 & 88.6 & 66.3 \\
FastVGGT \cite{shen2025fastvggt}  & 93.4 & 86.8 & 61.4 \\
\midrule
LiteVGGT & 93.3 & 86.6 & 60.6 \\
\bottomrule
\end{tabular}
}
\caption{Quantitative results of \textbf{camera pose estimation on the DTU dataset\cite{jensen2014large}.}}
\label{tab:pose_DTU}
\end{table}

\subsection{Pose Estimation}

\textbf{Quantitative comparisons.} 
As shown in Table \ref{tab:combined_dtu_co3d} (right) and Table \ref{tab:pose_DTU}, LiteVGGT achieves pose accuracy comparable to VGGT, even after applying extensive lossy operations to accelerate the original model. This highlights the effectiveness of our lightweight design. Finally, on large-scale outdoor scenes (Table \ref{tab:pos_tnt}), LiteVGGT surpasses FastVGGT in AUC@30 and remains close to VGGT, demonstrating strong pose estimation capability.

\textbf{Qualitative comparisons.}
As shown in the top part of Figure \ref{fig:main_result}, we visualize the predicted camera trajectories on a ScanNet-50 scene. In several scenes, the trajectories estimated by LiteVGGT align more closely with the ground truth, highlighting its accuracy in camera pose estimation.

\subsection{Efficiency}
Given that efficiency is a major contribution of LiteVGGT, we provide a focused analysis of its computational performance. VGGT runs out of memory when processing 500 images. VGGT$^*$ still requires over 20 minutes to process 1000 images on an H20 GPU, and increasing the input to 1100 images at 392×518 resolution causes out-of-memory even on the same 96 GiB device.

Building on VGGT$^*$, we introduce geometry-aware token merging, which reduces the number of tokens participating in global attention and delivers over a 4× latency reduction. Importantly, our geometry-aware strategy relies only on Sobel operators and average pooling, enabling the computation of the GA maps for 1000 images to finish in under one second.
To further improve efficiency, we cache the merge indices, reducing latency by about 25\%. Applying FP8 quantization brings an additional 33\% latency reduction and lowers memory consumption by roughly 25\%. All these optimizations are carefully designed to preserve accuracy while enabling the lightweight design of LiteVGGT.

\begin{table}[tp]
\centering
\setlength{\tabcolsep}{3pt}
\renewcommand{\arraystretch}{1.2}
\resizebox{1.0\linewidth}{!}{
\begin{tabular}{l|ccc|cc}
\toprule
\textbf{Methods} & \textbf{Acc.}$\downarrow$ & \textbf{Comp.}$\downarrow$ & \textbf{Overall}$\downarrow$ & \textbf{Time (s)} & \textbf{Mem. (GiB)} \\
\midrule
VGGT & 0.508 & 0.561 & 0.534 & 1275.6 & OOM \\
\cmidrule(lr){1-6}
+ GA Token Merging & 0.789 & 0.601 & 0.696 & 264.3 & 60.34 \\
+ Fine-tuning & 0.587 & 0.687 & 0.642 & 264.3 & 60.34 \\
+ Cache Merge Indices & 0.621 & 0.755 & 0.688 & 200.6 & 59.28 \\
+ FP8 Quantization & \textbf{0.652 }& \textbf{0.780} & \textbf{0.716} & \textbf{127.9} & \textbf{45.31} \\
\bottomrule
\end{tabular}
}
\caption{Ablation study on the each propsoed modules of LiteVGGT. (See more details in Supp. file)}
\label{tab:efficiency_anay}
\end{table}

\subsection{Ablation Studies}

We conduct ablation studies to quantify the contribution of each optimization step in LiteVGGT, as shown in Table~\ref{tab:efficiency_anay}. Accuracy is evaluated using DTU reconstruction, and efficiency is measured by the latency and memory usage when processing 1000 images at 392×518 resolution on an H20 GPU. VGGT runs out of memory at this scale, while VGGT$^*$ requires over 20 minutes.

\textbf{(1) Geometry-aware token merging} reduces the number of global-attention tokens, yielding over a 4× latency reduction and substantial memory savings with almost no loss in reconstruction quality.  

\textbf{(2) Caching merge indices} eliminates repeated large-matrix computations and reduces latency by 25\% with only a minor accuracy change.  

\textbf{(3) Fine-tuning} recovers the accuracy drop introduced by GA token merging.  

\textbf{(4) FP8 quantization} provides an additional 33\% latency reduction and 25\% memory savings while maintaining accuracy.

\section{Conclusions}


LiteVGGT offers an efficient and scalable alternative to existing 3D vision foundation models by substantially lowering their computational and memory requirements. Our study shows that large-scale scene reconstruction does not necessarily require heavy global attention computation, and that carefully designed token reduction strategies are sufficient to preserve high geometric fidelity. Through systematic optimization and evaluation, LiteVGGT achieves strong reconstruction and pose estimation accuracy while enabling practical processing of long image sequences that exceed the limits of VGGT. These results highlight the potential of lightweight architectures for real-world, large-scale 3D applications. Future work will explore extending this framework to video inputs and even larger and more complex scenes.

{
    \small
    \bibliographystyle{ieeenat_fullname}
    \bibliography{main}

\begin{thebibliography}{60}
\providecommand{\natexlab}[1]{#1}
\providecommand{\url}[1]{\texttt{#1}}
\expandafter\ifx\csname urlstyle\endcsname\relax
  \providecommand{\doi}[1]{doi: #1}\else
  \providecommand{\doi}{doi: \begingroup \urlstyle{rm}\Url}\fi

\bibitem[Azinovi{\'c} et~al.(2022)Azinovi{\'c}, Martin-Brualla, Goldman, Nie{\ss}ner, and Thies]{azinovic2022neural}
Dejan Azinovi{\'c}, Ricardo Martin-Brualla, Dan~B Goldman, Matthias Nie{\ss}ner, and Justus Thies.
\newblock Neural rgb-d surface reconstruction.
\newblock In \emph{Proceedings of the IEEE/CVF Conference on Computer Vision and Pattern Recognition}, pages 6290--6301, 2022.

\bibitem[Bhalgat et~al.(2023)Bhalgat, Laina, Henriques, Zisserman, and Vedaldi]{bhalgat23contrastive}
Yash~Sanjay Bhalgat, Iro Laina, Joao~F. Henriques, Andrew Zisserman, and Andrea Vedaldi.
\newblock Contrastive lift: 3d object instance segmentation by slow-fast contrastive fusion.
\newblock In \emph{Proceedings of Advances in Neural Information Processing Systems (NeurIPS)}, 2023.

\bibitem[Bolya and Hoffman(2023)]{bolya2023token}
Daniel Bolya and Judy Hoffman.
\newblock Token merging for fast stable diffusion.
\newblock In \emph{Proceedings of the IEEE/CVF conference on computer vision and pattern recognition}, pages 4599--4603, 2023.

\bibitem[Bolya et~al.(2022)Bolya, Fu, Dai, Zhang, Feichtenhofer, and Hoffman]{bolya2022token}
Daniel Bolya, Cheng-Yang Fu, Xiaoliang Dai, Peizhao Zhang, Christoph Feichtenhofer, and Judy Hoffman.
\newblock Token merging: Your vit but faster.
\newblock \emph{arXiv preprint arXiv:2210.09461}, 2022.

\bibitem[Cabon et~al.(2020)Cabon, Murray, and Humenberger]{cabon2020virtual}
Yohann Cabon, Naila Murray, and Martin Humenberger.
\newblock Virtual kitti 2.
\newblock \emph{arXiv preprint arXiv:2001.10773}, 2020.

\bibitem[Cao et~al.(2022{\natexlab{a}})Cao, Ren, and Fu]{cao2022mvsformer}
Chenjie Cao, Xinlin Ren, and Yanwei Fu.
\newblock Mvsformer: Multi-view stereo by learning robust image features and temperature-based depth.
\newblock \emph{arXiv preprint arXiv:2208.02541}, 2022{\natexlab{a}}.

\bibitem[Cao et~al.(2022{\natexlab{b}})Cao, Wang, Chemerys, Shakhrai, Hu, Fu, Makoviichuk, Tulyakov, and Ren]{cao2022mobiler2l}
Junli Cao, Huan Wang, Pavlo Chemerys, Vladislav Shakhrai, Ju Hu, Yun Fu, Denys Makoviichuk, Sergey Tulyakov, and Jian Ren.
\newblock Real-time neural light field on mobile devices.
\newblock \emph{arXiv preprint arXiv:2212.08057}, 2022{\natexlab{b}}.

\bibitem[Cao et~al.(2023)Cao, Paranjape, and Hajishirzi]{cao2023pumer}
Qingqing Cao, Bhargavi Paranjape, and Hannaneh Hajishirzi.
\newblock Pumer: Pruning and merging tokens for efficient vision language models.
\newblock \emph{arXiv preprint arXiv:2305.17530}, 2023.

\bibitem[Cheng et~al.(2024)Cheng, Yin, Wang, Chen, Wang, and Yang]{cheng2024adaptive}
Junda Cheng, Wei Yin, Kaixuan Wang, Xiaozhi Chen, Shijie Wang, and Xin Yang.
\newblock Adaptive fusion of single-view and multi-view depth for autonomous driving.
\newblock In \emph{Proceedings of the IEEE/CVF Conference on Computer Vision and Pattern Recognition}, pages 10138--10147, 2024.

\bibitem[Cheng et~al.(2022)Cheng, Chen, Yin, Xu, and Chen]{cheng2022exploiting}
Kai Cheng, Hao Chen, Wei Yin, Guangkai Xu, and Xuejin Chen.
\newblock Exploiting correspondences with all-pairs correlations for multi-view depth estimation.
\newblock \emph{arXiv preprint arXiv:2205.02481}, 2022.

\bibitem[Dai et~al.(2017)Dai, Chang, Savva, Halber, Funkhouser, and Nie{\ss}ner]{dai2017scannet}
Angela Dai, Angel~X Chang, Manolis Savva, Maciej Halber, Thomas Funkhouser, and Matthias Nie{\ss}ner.
\newblock Scannet: Richly-annotated 3d reconstructions of indoor scenes.
\newblock In \emph{Proceedings of the IEEE conference on computer vision and pattern recognition}, pages 5828--5839, 2017.

\bibitem[Dosovitskiy(2020)]{dosovitskiy2020image}
Alexey Dosovitskiy.
\newblock An image is worth 16x16 words: Transformers for image recognition at scale.
\newblock \emph{arXiv preprint arXiv:2010.11929}, 2020.

\bibitem[Gu et~al.(2020)Gu, Fan, Zhu, Dai, Tan, and Tan]{gu2020cascade}
Xiaodong Gu, Zhiwen Fan, Siyu Zhu, Zuozhuo Dai, Feitong Tan, and Ping Tan.
\newblock Cascade cost volume for high-resolution multi-view stereo and stereo matching.
\newblock In \emph{Proceedings of the IEEE/CVF conference on computer vision and pattern recognition}, pages 2495--2504, 2020.

\bibitem[Huang et~al.(2018)Huang, Matzen, Kopf, Ahuja, and Huang]{huang2018deepmvs}
Po-Han Huang, Kevin Matzen, Johannes Kopf, Narendra Ahuja, and Jia-Bin Huang.
\newblock Deepmvs: Learning multi-view stereopsis.
\newblock In \emph{Proceedings of the IEEE conference on computer vision and pattern recognition}, pages 2821--2830, 2018.

\bibitem[Jensen et~al.(2014)Jensen, Dahl, Vogiatzis, Tola, and Aan{\ae}s]{jensen2014large}
Rasmus Jensen, Anders Dahl, George Vogiatzis, Engin Tola, and Henrik Aan{\ae}s.
\newblock Large scale multi-view stereopsis evaluation.
\newblock In \emph{Proceedings of the IEEE conference on computer vision and pattern recognition}, pages 406--413, 2014.

\bibitem[Knapitsch et~al.(2017)Knapitsch, Park, Zhou, and Koltun]{knapitsch2017tanks}
Arno Knapitsch, Jaesik Park, Qian-Yi Zhou, and Vladlen Koltun.
\newblock Tanks and temples: Benchmarking large-scale scene reconstruction.
\newblock \emph{ACM Transactions on Graphics (ToG)}, 36\penalty0 (4):\penalty0 1--13, 2017.

\bibitem[Li and Snavely(2018)]{li2018megadepth}
Zhengqi Li and Noah Snavely.
\newblock Megadepth: Learning single-view depth prediction from internet photos.
\newblock In \emph{Proceedings of the IEEE conference on computer vision and pattern recognition}, pages 2041--2050, 2018.

\bibitem[Lin and Shou(2025)]{lin2025vlog}
Kevin~Qinghong Lin and Mike~Zheng Shou.
\newblock Vlog: Video-language models by generative retrieval of narration vocabulary, 2025.

\bibitem[Ling et~al.(2024)Ling, Sheng, Tu, Zhao, Xin, Wan, Yu, Guo, Yu, Lu, et~al.]{ling2024dl3dv}
Lu Ling, Yichen Sheng, Zhi Tu, Wentian Zhao, Cheng Xin, Kun Wan, Lantao Yu, Qianyu Guo, Zixun Yu, Yawen Lu, et~al.
\newblock Dl3dv-10k: A large-scale scene dataset for deep learning-based 3d vision.
\newblock In \emph{Proceedings of the IEEE/CVF Conference on Computer Vision and Pattern Recognition}, pages 22160--22169, 2024.

\bibitem[Long et~al.(2024)Long, Guo, Lin, Liu, Dou, Liu, Ma, Zhang, Habermann, Theobalt, et~al.]{long2024wonder3d}
Xiaoxiao Long, Yuan-Chen Guo, Cheng Lin, Yuan Liu, Zhiyang Dou, Lingjie Liu, Yuexin Ma, Song-Hai Zhang, Marc Habermann, Christian Theobalt, et~al.
\newblock Wonder3d: Single image to 3d using cross-domain diffusion.
\newblock In \emph{Proceedings of the IEEE/CVF conference on computer vision and pattern recognition}, pages 9970--9980, 2024.

\bibitem[Lu et~al.(2025)Lu, Huang, Li, Dou, Lin, Cui, Dong, Yeung, Wang, and Liu]{lu2025align3r}
Jiahao Lu, Tianyu Huang, Peng Li, Zhiyang Dou, Cheng Lin, Zhiming Cui, Zhen Dong, Sai-Kit Yeung, Wenping Wang, and Yuan Liu.
\newblock Align3r: Aligned monocular depth estimation for dynamic videos.
\newblock In \emph{Proceedings of the Computer Vision and Pattern Recognition Conference}, pages 22820--22830, 2025.

\bibitem[Mildenhall et~al.(2021)Mildenhall, Srinivasan, Tancik, Barron, Ramamoorthi, and Ng]{mildenhall2021nerf}
Ben Mildenhall, Pratul~P Srinivasan, Matthew Tancik, Jonathan~T Barron, Ravi Ramamoorthi, and Ren Ng.
\newblock Nerf: Representing scenes as neural radiance fields for view synthesis.
\newblock \emph{Communications of the ACM}, 65\penalty0 (1):\penalty0 99--106, 2021.

\bibitem[Mur-Artal et~al.(2015)Mur-Artal, Montiel, and Tardos]{mur2015orb}
Raul Mur-Artal, Jose Maria~Martinez Montiel, and Juan~D Tardos.
\newblock Orb-slam: A versatile and accurate monocular slam system.
\newblock \emph{IEEE transactions on robotics}, 31\penalty0 (5):\penalty0 1147--1163, 2015.

\bibitem[Murai et~al.(2025)Murai, Dexheimer, and Davison]{murai2025mast3r}
Riku Murai, Eric Dexheimer, and Andrew~J Davison.
\newblock Mast3r-slam: Real-time dense slam with 3d reconstruction priors.
\newblock In \emph{Proceedings of the Computer Vision and Pattern Recognition Conference}, pages 16695--16705, 2025.

\bibitem[Nawrot et~al.(2023)Nawrot, Chorowski, Lancucki, and Ponti]{nawrot2023efficient}
Piotr Nawrot, Jan Chorowski, Adrian Lancucki, and Edoardo~Maria Ponti.
\newblock Efficient transformers with dynamic token pooling.
\newblock In \emph{Proceedings of the 61st Annual Meeting of the Association for Computational Linguistics (Volume 1: Long Papers)}, pages 6403--6417, 2023.

\bibitem[Neuhold et~al.(2017)Neuhold, Ollmann, Rota~Bulo, and Kontschieder]{neuhold2017mapillary}
Gerhard Neuhold, Tobias Ollmann, Samuel Rota~Bulo, and Peter Kontschieder.
\newblock The mapillary vistas dataset for semantic understanding of street scenes.
\newblock In \emph{Proceedings of the IEEE international conference on computer vision}, pages 4990--4999, 2017.

\bibitem[Opt and Opt()]{optquantized}
Speed Opt and Memory Opt.
\newblock Quantized visual geometry grounded transformer.

\bibitem[Oquab et~al.(2023)Oquab, Darcet, Moutakanni, Vo, Szafraniec, Khalidov, Fernandez, Haziza, Massa, El-Nouby, et~al.]{oquab2023dinov2}
Maxime Oquab, Timoth{\'e}e Darcet, Th{\'e}o Moutakanni, Huy Vo, Marc Szafraniec, Vasil Khalidov, Pierre Fernandez, Daniel Haziza, Francisco Massa, Alaaeldin El-Nouby, et~al.
\newblock Dinov2: Learning robust visual features without supervision.
\newblock \emph{arXiv preprint arXiv:2304.07193}, 2023.

\bibitem[Pan et~al.(2023)Pan, Charron, Yang, Peters, Whelan, Kong, Parkhi, Newcombe, and Ren]{pan2023aria}
Xiaqing Pan, Nicholas Charron, Yongqian Yang, Scott Peters, Thomas Whelan, Chen Kong, Omkar Parkhi, Richard Newcombe, and Yuheng~Carl Ren.
\newblock Aria digital twin: A new benchmark dataset for egocentric 3d machine perception.
\newblock In \emph{Proceedings of the IEEE/CVF International Conference on Computer Vision}, pages 20133--20143, 2023.

\bibitem[Park et~al.(2020)Park, Sinha, Barron, Bouaziz, Goldman, Seitz, and Martin-Brualla]{Park20arxiv_nerfies}
Keunhong Park, Utkarsh Sinha, Jonathan~T. Barron, Sofien Bouaziz, Dan Goldman, Steven Seitz, and Ricardo Martin-Brualla.
\newblock Deformable neural radiance fields.
\newblock \emph{https://arxiv.org/abs/2011.12948}, 2020.

\bibitem[Pietruszka et~al.(2022)Pietruszka, Borchmann, and Garncarek]{pietruszka2022sparsifying}
Micha{\l} Pietruszka, {\L}ukasz Borchmann, and {\L}ukasz Garncarek.
\newblock Sparsifying transformer models with trainable representation pooling.
\newblock In \emph{Proceedings of the 60th Annual Meeting of the Association for Computational Linguistics (Volume 1: Long Papers)}, pages 8616--8633, 2022.

\bibitem[Reizenstein et~al.(2021)Reizenstein, Shapovalov, Henzler, Sbordone, Labatut, and Novotny]{reizenstein2021common}
Jeremy Reizenstein, Roman Shapovalov, Philipp Henzler, Luca Sbordone, Patrick Labatut, and David Novotny.
\newblock Common objects in 3d: Large-scale learning and evaluation of real-life 3d category reconstruction.
\newblock In \emph{Proceedings of the IEEE/CVF international conference on computer vision}, pages 10901--10911, 2021.

\bibitem[Roberts et~al.(2021)Roberts, Ramapuram, Ranjan, Kumar, Bautista, Paczan, Webb, and Susskind]{roberts2021hypersim}
Mike Roberts, Jason Ramapuram, Anurag Ranjan, Atulit Kumar, Miguel~Angel Bautista, Nathan Paczan, Russ Webb, and Joshua~M Susskind.
\newblock Hypersim: A photorealistic synthetic dataset for holistic indoor scene understanding.
\newblock In \emph{Proceedings of the IEEE/CVF international conference on computer vision}, pages 10912--10922, 2021.

\bibitem[Rombach et~al.(2021)Rombach, Blattmann, Lorenz, Esser, and Ommer]{rombach2021highresolution}
Robin Rombach, Andreas Blattmann, Dominik Lorenz, Patrick Esser, and Björn Ommer.
\newblock High-resolution image synthesis with latent diffusion models, 2021.

\bibitem[Ryoo et~al.(2021)Ryoo, Piergiovanni, Arnab, Dehghani, and Angelova]{ryoo2021tokenlearner}
Michael Ryoo, AJ Piergiovanni, Anurag Arnab, Mostafa Dehghani, and Anelia Angelova.
\newblock Tokenlearner: Adaptive space-time tokenization for videos.
\newblock \emph{Advances in neural information processing systems}, 34:\penalty0 12786--12797, 2021.

\bibitem[Shen et~al.(2025)Shen, Zhang, Qu, and Cao]{shen2025fastvggt}
You Shen, Zhipeng Zhang, Yansong Qu, and Liujuan Cao.
\newblock Fastvggt: Training-free acceleration of visual geometry transformer.
\newblock \emph{arXiv preprint arXiv:2509.02560}, 2025.

\bibitem[Shotton et~al.(2013)Shotton, Glocker, Zach, Izadi, Criminisi, and Fitzgibbon]{shotton2013scene}
Jamie Shotton, Ben Glocker, Christopher Zach, Shahram Izadi, Antonio Criminisi, and Andrew Fitzgibbon.
\newblock Scene coordinate regression forests for camera relocalization in rgb-d images.
\newblock In \emph{Proceedings of the IEEE conference on computer vision and pattern recognition}, pages 2930--2937, 2013.

\bibitem[Smart et~al.(2024)Smart, Zheng, Laina, and Prisacariu]{smart2024splatt3r}
Brandon Smart, Chuanxia Zheng, Iro Laina, and Victor~Adrian Prisacariu.
\newblock Splatt3r: Zero-shot gaussian splatting from uncalibrated image pairs.
\newblock \emph{arXiv preprint arXiv:2408.13912}, 2024.

\bibitem[Song et~al.(2024)Song, Chai, Wang, Zhang, Zhou, Wu, Chi, Guo, Ye, Zhang, et~al.]{song2024moviechat}
Enxin Song, Wenhao Chai, Guanhong Wang, Yucheng Zhang, Haoyang Zhou, Feiyang Wu, Haozhe Chi, Xun Guo, Tian Ye, Yanting Zhang, et~al.
\newblock Moviechat: From dense token to sparse memory for long video understanding.
\newblock In \emph{Proceedings of the IEEE/CVF Conference on Computer Vision and Pattern Recognition}, pages 18221--18232, 2024.

\bibitem[Sucar et~al.(2021)Sucar, Liu, Ortiz, and Davison]{sucar2021imap}
Edgar Sucar, Shikun Liu, Joseph Ortiz, and Andrew~J Davison.
\newblock imap: Implicit mapping and positioning in real-time.
\newblock In \emph{Proceedings of the IEEE/CVF international conference on computer vision}, pages 6229--6238, 2021.

\bibitem[Szymanowicz et~al.(2025)Szymanowicz, Insafutdinov, Zheng, Campbell, Henriques, Rupprecht, and Vedaldi]{szymanowicz2025flash3d}
Stanislaw Szymanowicz, Eldar Insafutdinov, Chuanxia Zheng, Dylan Campbell, Joao~F Henriques, Christian Rupprecht, and Andrea Vedaldi.
\newblock Flash3d: Feed-forward generalisable 3d scene reconstruction from a single image.
\newblock In \emph{2025 International Conference on 3D Vision (3DV)}, pages 670--681. IEEE, 2025.

\bibitem[Takikawa et~al.(2022)Takikawa, Evans, Tremblay, M{\"u}ller, McGuire, Jacobson, and Fidler]{takikawa2022variable}
Towaki Takikawa, Alex Evans, Jonathan Tremblay, Thomas M{\"u}ller, Morgan McGuire, Alec Jacobson, and Sanja Fidler.
\newblock Variable bitrate neural fields.
\newblock In \emph{ACM SIGGRAPH 2022 Conference Proceedings}, pages 1--9, 2022.

\bibitem[Thrun(2002)]{thrun2002probabilistic}
Sebastian Thrun.
\newblock Probabilistic robotics.
\newblock \emph{Communications of the ACM}, 45\penalty0 (3):\penalty0 52--57, 2002.

\bibitem[Wang et~al.(2025{\natexlab{a}})Wang, Chen, Karaev, Vedaldi, Rupprecht, and Novotny]{wang2025vggt}
Jianyuan Wang, Minghao Chen, Nikita Karaev, Andrea Vedaldi, Christian Rupprecht, and David Novotny.
\newblock Vggt: Visual geometry grounded transformer.
\newblock In \emph{Proceedings of the Computer Vision and Pattern Recognition Conference}, pages 5294--5306, 2025{\natexlab{a}}.

\bibitem[Wang et~al.(2025{\natexlab{b}})Wang, Zhang, Holynski, Efros, and Kanazawa]{wang2025continuous}
Qianqian Wang, Yifei Zhang, Aleksander Holynski, Alexei~A Efros, and Angjoo Kanazawa.
\newblock Continuous 3d perception model with persistent state.
\newblock In \emph{Proceedings of the Computer Vision and Pattern Recognition Conference}, pages 10510--10522, 2025{\natexlab{b}}.

\bibitem[Wang et~al.(2022)Wang, Li, and Dai]{wang2022efficient}
Shaoqian Wang, Bo Li, and Yuchao Dai.
\newblock Efficient multi-view stereo by iterative dynamic cost volume.
\newblock In \emph{Proceedings of the IEEE/CVF conference on computer vision and pattern recognition}, pages 8655--8664, 2022.

\bibitem[Wang et~al.(2024)Wang, Leroy, Cabon, Chidlovskii, and Revaud]{wang2024dust3r}
Shuzhe Wang, Vincent Leroy, Yohann Cabon, Boris Chidlovskii, and Jerome Revaud.
\newblock Dust3r: Geometric 3d vision made easy.
\newblock In \emph{Proceedings of the IEEE/CVF Conference on Computer Vision and Pattern Recognition}, pages 20697--20709, 2024.

\bibitem[Xia et~al.(2024)Xia, Fu, Liu, and Wang]{xia2024rgbd}
Hongchi Xia, Yang Fu, Sifei Liu, and Xiaolong Wang.
\newblock Rgbd objects in the wild: Scaling real-world 3d object learning from rgb-d videos.
\newblock In \emph{Proceedings of the IEEE/CVF Conference on Computer Vision and Pattern Recognition}, pages 22378--22389, 2024.

\bibitem[Xu et~al.(2022)Xu, Kong, Tao, and Pollefeys]{xu2022multi}
Qingshan Xu, Weihang Kong, Wenbing Tao, and Marc Pollefeys.
\newblock Multi-scale geometric consistency guided and planar prior assisted multi-view stereo.
\newblock \emph{IEEE Transactions on Pattern Analysis and Machine Intelligence}, 45\penalty0 (4):\penalty0 4945--4963, 2022.

\bibitem[Yang et~al.(2025)Yang, Sax, Liang, Henaff, Tang, Cao, Chai, Meier, and Feiszli]{yang2025fast3r}
Jianing Yang, Alexander Sax, Kevin~J Liang, Mikael Henaff, Hao Tang, Ang Cao, Joyce Chai, Franziska Meier, and Matt Feiszli.
\newblock Fast3r: Towards 3d reconstruction of 1000+ images in one forward pass.
\newblock In \emph{Proceedings of the Computer Vision and Pattern Recognition Conference}, pages 21924--21935, 2025.

\bibitem[Yang et~al.(2024)Yang, Kang, Huang, Zhao, Xu, Feng, and Zhao]{yang2024depth}
Lihe Yang, Bingyi Kang, Zilong Huang, Zhen Zhao, Xiaogang Xu, Jiashi Feng, and Hengshuang Zhao.
\newblock Depth anything v2.
\newblock \emph{Advances in Neural Information Processing Systems}, 37:\penalty0 21875--21911, 2024.

\bibitem[Yao et~al.(2018)Yao, Luo, Li, Fang, and Quan]{yao2018mvsnet}
Yao Yao, Zixin Luo, Shiwei Li, Tian Fang, and Long Quan.
\newblock Mvsnet: Depth inference for unstructured multi-view stereo.
\newblock In \emph{Proceedings of the European conference on computer vision (ECCV)}, pages 767--783, 2018.

\bibitem[Yao et~al.(2020)Yao, Luo, Li, Zhang, Ren, Zhou, Fang, and Quan]{yao2020blendedmvs}
Yao Yao, Zixin Luo, Shiwei Li, Jingyang Zhang, Yufan Ren, Lei Zhou, Tian Fang, and Long Quan.
\newblock Blendedmvs: A large-scale dataset for generalized multi-view stereo networks.
\newblock In \emph{Proceedings of the IEEE/CVF conference on computer vision and pattern recognition}, pages 1790--1799, 2020.

\bibitem[Yeshwanth et~al.(2023)Yeshwanth, Liu, Nie{\ss}ner, and Dai]{yeshwanth2023scannet++}
Chandan Yeshwanth, Yueh-Cheng Liu, Matthias Nie{\ss}ner, and Angela Dai.
\newblock Scannet++: A high-fidelity dataset of 3d indoor scenes.
\newblock In \emph{Proceedings of the IEEE/CVF International Conference on Computer Vision}, pages 12--22, 2023.

\bibitem[Zhang et~al.(2023)Zhang, Li, and Bing]{zhang2023video}
Hang Zhang, Xin Li, and Lidong Bing.
\newblock Video-llama: An instruction-tuned audio-visual language model for video understanding.
\newblock \emph{arXiv preprint arXiv:2306.02858}, 2023.

\bibitem[Zhang et~al.(2024{\natexlab{a}})Zhang, Herrmann, Hur, Jampani, Darrell, Cole, Sun, and Yang]{zhang2024monst3r}
Junyi Zhang, Charles Herrmann, Junhwa Hur, Varun Jampani, Trevor Darrell, Forrester Cole, Deqing Sun, and Ming-Hsuan Yang.
\newblock Monst3r: A simple approach for estimating geometry in the presence of motion.
\newblock \emph{arXiv preprint arXiv:2410.03825}, 2024{\natexlab{a}}.

\bibitem[Zhang et~al.(2020)Zhang, Riegler, Snavely, and Koltun]{Zhang20arxiv_nerf++}
Kai Zhang, Gernot Riegler, Noah Snavely, and Vladlen Koltun.
\newblock {NERF++}: Analyzing and improving neural radiance fields.
\newblock \emph{https://arxiv.org/abs/2010.07492}, 2020.

\bibitem[Zhang et~al.(2024{\natexlab{b}})Zhang, Bi, Tan, Xiangli, Zhao, Sunkavalli, and Xu]{zhang2024gs}
Kai Zhang, Sai Bi, Hao Tan, Yuanbo Xiangli, Nanxuan Zhao, Kalyan Sunkavalli, and Zexiang Xu.
\newblock Gs-lrm: Large reconstruction model for 3d gaussian splatting.
\newblock In \emph{European Conference on Computer Vision}, pages 1--19. Springer, 2024{\natexlab{b}}.

\bibitem[Zhao et~al.(2023)Zhao, Misra, Kr{\"a}henb{\"u}hl, and Girdhar]{zhao2023learning}
Yue Zhao, Ishan Misra, Philipp Kr{\"a}henb{\"u}hl, and Rohit Girdhar.
\newblock Learning video representations from large language models.
\newblock In \emph{Proceedings of the IEEE/CVF Conference on Computer Vision and Pattern Recognition}, pages 6586--6597, 2023.

\bibitem[Zhuo et~al.(2025)Zhuo, Zheng, Guo, Wu, Zhou, and Lu]{zhuo2025streaming}
Dong Zhuo, Wenzhao Zheng, Jiahe Guo, Yuqi Wu, Jie Zhou, and Jiwen Lu.
\newblock Streaming 4d visual geometry transformer.
\newblock \emph{arXiv preprint arXiv:2507.11539}, 2025.

\end{thebibliography}
}

\clearpage
\setcounter{page}{1}
\maketitlesupplementary

\section{More Implementation Details}

\paragraph{Fine-tuning Details.}
Following VGGT's official fine-tuning pipeline, we fine-tune only the aggregator together with the camera head and depth head, starting from the pretrained VGGT checkpoints. 
We use a diverse mixture of synthetic and real-world datasets, including Co3Dv2\cite{reizenstein2021common} , BlendMVS\cite{yao2020blendedmvs}, DL3DV\cite{ling2024dl3dv}, MegaDepth\cite{li2018megadepth}, WildRGB\cite{xia2024rgbd}, ScanNet++\cite{yeshwanth2023scannet++}, HyperSim\cite{roberts2021hypersim}, Mapillary\cite{neuhold2017mapillary}, Replica\cite{sucar2021imap}, MVS-Synth\cite{huang2018deepmvs}, Virtual KITTI\cite{cabon2020virtual}, Aria Synthetic Environments, and Aria Digital Twin\cite{pan2023aria}.
We sample 4--48 images per batch and train for 20K iterations on 8 H20 GPUs (approximately 3 days). 
To stabilize training under token merging, we adopt a composite learning rate schedule: a 5\% linear warm-up from \(1\times 10^{-6}\) to \(4\times 10^{-5}\), followed by cosine decay to \(7\times 10^{-7}\) over the remaining iterations.

\paragraph{FP8 Quantization Details.}
We apply FP8 mixed precision via NVIDIA's Transformer Engine directly during inference. 
To balance efficiency and accuracy, we quantize only the modules inside the aggregator (including the DINOv2 encoder). 
For example, each \texttt{nn.LayerNorm} + two-layer MLP block is replaced with its FP8-enabled \texttt{te.LayerNormMLP} counterpart, while the prediction heads remain in bfloat16 since quantizing them leads to a noticeable drop in accuracy. 
We adopt an FP8 (E4M3) delayed-scaling recipe with an 80-step amax history and max-based amax computation.

\section{Additional Ablation Studies}

\paragraph{Caching merge indices.}
Leveraging the stability of inter-layer token similarity, we cache merge indices and evaluate how different recomputation intervals affect the accuracy–efficiency trade-off. 

\begin{table}[h]
\centering
\setlength{\tabcolsep}{5pt}
\renewcommand{\arraystretch}{1.15}
\resizebox{1.0\linewidth}{!}{
\begin{tabular}{c|c|c|ccc|c}
\toprule
\textbf{Interval (layers)} 
& \textbf{Total Recompute} 
& \textbf{CD} $\downarrow$ 
& \textbf{Acc} $\downarrow$ 
& \textbf{Comp} $\downarrow$ 
& \textbf{Overall} $\downarrow$ 
& \textbf{Time (s)} $\downarrow$ \\
\midrule
1  & 24 & 0.421 & 0.691 & 0.800 & 0.745  & 265.1 \\
2  & 12 & 0.440 & 0.701 & 0.778 & 0.741 & 230.5 \\
3  &  8 & 0.452 & 0.721 & 0.788 & 0.754 &  214.1 \\
6  &  4 & 0.467 & 0.746 & 0.775 & 0.761 & 202.0 \\
24 &  1 & 0.555 & 3.761 & 5.342 & 4.552 & 193.1 \\
\bottomrule
\end{tabular}
}
\caption{\textbf{Ablation on merge-indices recomputation intervals.}
Quantitative results of point cloud reconstruction on the Scannet-50\cite{dai2017scannet} and DTU\cite{jensen2014large} dataset.
}
\label{tab:caching_merge_indices}
\end{table}

As shown in Table~\ref{tab:caching_merge_indices}, enlarging the interval from every 1 layer to every 6 layers preserves nearly the same point cloud reconstruction quality while substantially reducing inference latency, further validating the stability of token similarity across adjacent layers. In contrast, using an excessively large interval (i.e., computing only at the first layer) causes a clear drop in reconstruction accuracy. Based on this trade-off, we adopt the 6-layer interval (4 total computations) as our preferred setting.

\paragraph{Geometry-aware Token Merging.}
We further evaluate our Geometry-aware Token Merging against standard similarity-based merging. As shown in Table \ref{tab:ga_merge_ablation}, our method better preserves crucial geometric information and achieves better performance on both datasets.
Meanwhile, the latency remains almost unchanged, highlighting the efficiency of our approach.

\begin{table}[h]
\centering
\setlength{\tabcolsep}{5pt}
\renewcommand{\arraystretch}{1.15}
\resizebox{1.0\linewidth}{!}{
\begin{tabular}{l|c|ccc|c}
\toprule
\textbf{Method} 
& \textbf{CD} $\downarrow$
& \textbf{Acc} $\downarrow$
& \textbf{Comp} $\downarrow$
& \textbf{Overall} $\downarrow$
& \textbf{Time (s)} $\downarrow$ \\
\midrule
VGGT$^*$ & 0.485 & 0.508 & 0.561 & 0.534 &  1275.0 \\
VGGT$^*$+Naive Token Merging & 0.442 & 0.824 & 0.655 & 0.739 & 202.0 \\
\textbf{VGGT$^*$+GA Token Merging (Ours)} & 0.402 & 0.789 & 0.601 & 0.696 & 202.4 \\
\bottomrule
\end{tabular}
}
\caption{\textbf{Ablation on Geometry-aware Token Merging.}
Quantitative results of point cloud reconstruction on the Scannet-50\cite{dai2017scannet} and DTU\cite{jensen2014large} dataset.
}
\label{tab:ga_merge_ablation}
\end{table}

\section{Robotic Grasping Experiments}
\begin{figure}[h]
\centering
\includegraphics[width=1\linewidth]{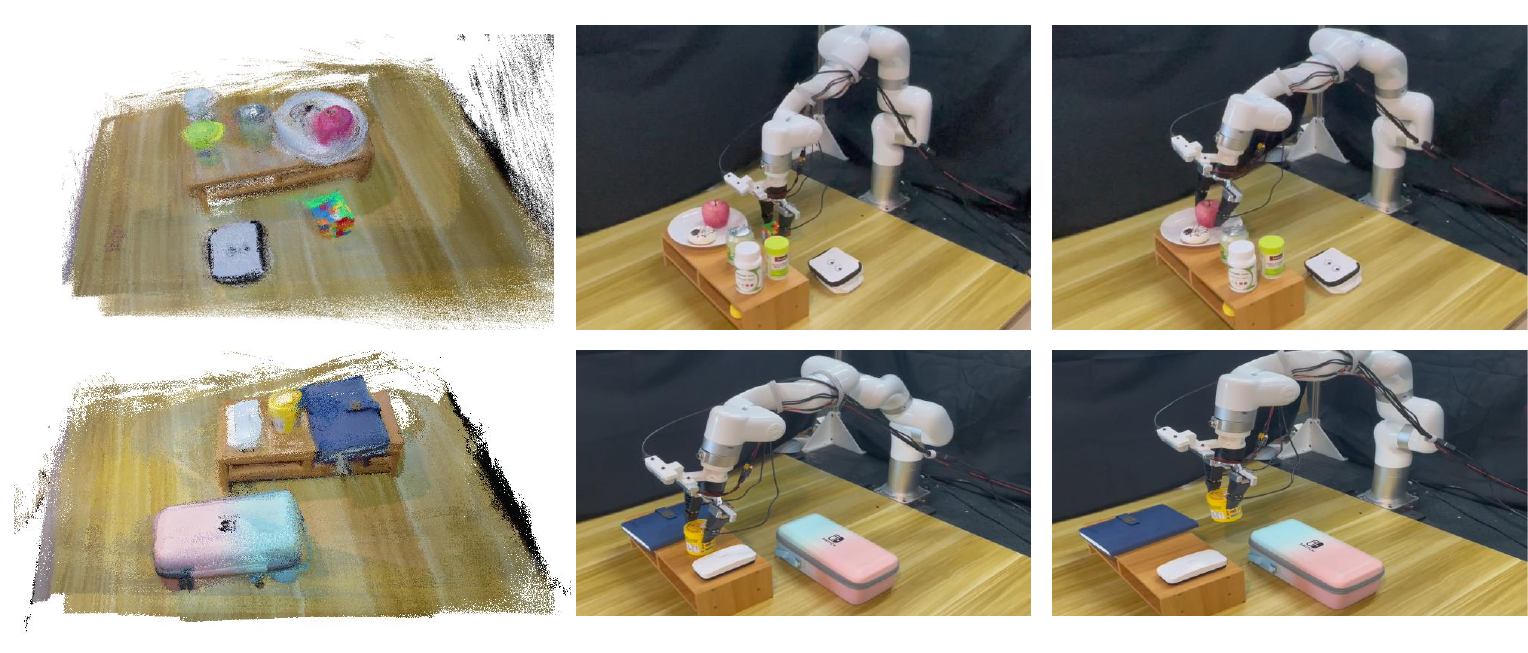}
\vspace{-5mm}
\caption{Robotic grasping demonstration. Left: reconstructed point clouds from LiteVGGT. Right: corresponding grasp execution snapshots. The full videos are provided in the supplementary material (video.mp4).}
\label{fig:demo}
\end{figure}

To further verify that LiteVGGT remains practical even with its 10× reconstruction speed-up, we conduct robotic grasping experiments. As shown in Fig. \ref{fig:demo}, LiteVGGT reconstructs two real-world scenes, and the resulting point clouds are used for robotic arm grasping. Despite minor reconstruction deviations, the accuracy is sufficient for end-side grasp execution, demonstrating the practical reliability of LiteVGGT.

\section{Additional Visualizations}
\begin{figure*}[t]
\centering
\includegraphics[width=1\linewidth]{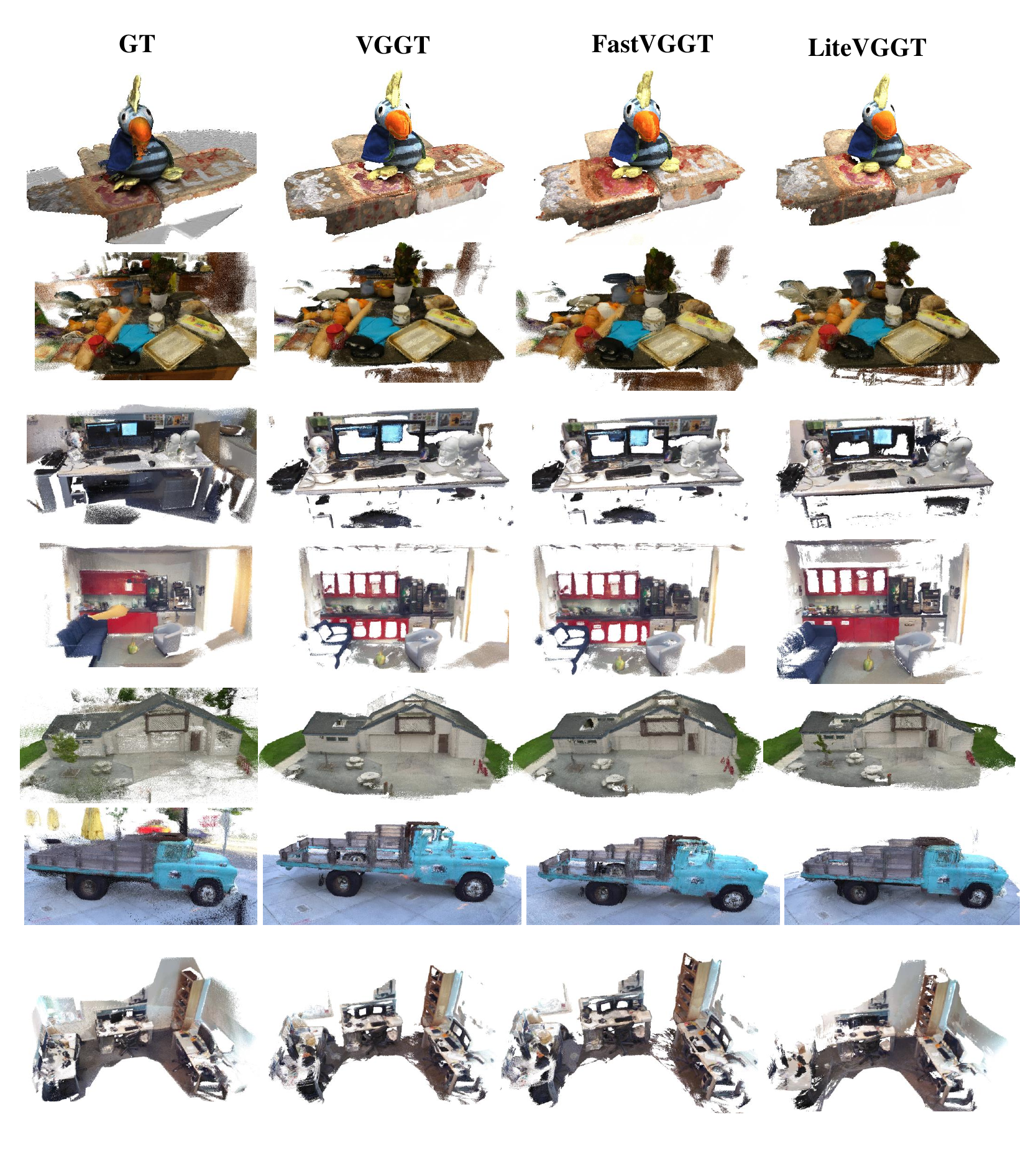}
\vspace{-5mm}
\caption{Additional 3D reconstruction visualizations from small indoor objects to large outdoor scenes.}
\label{fig:scene}
\end{figure*}

\begin{figure*}[t]
\centering
\includegraphics[width=0.9\linewidth]{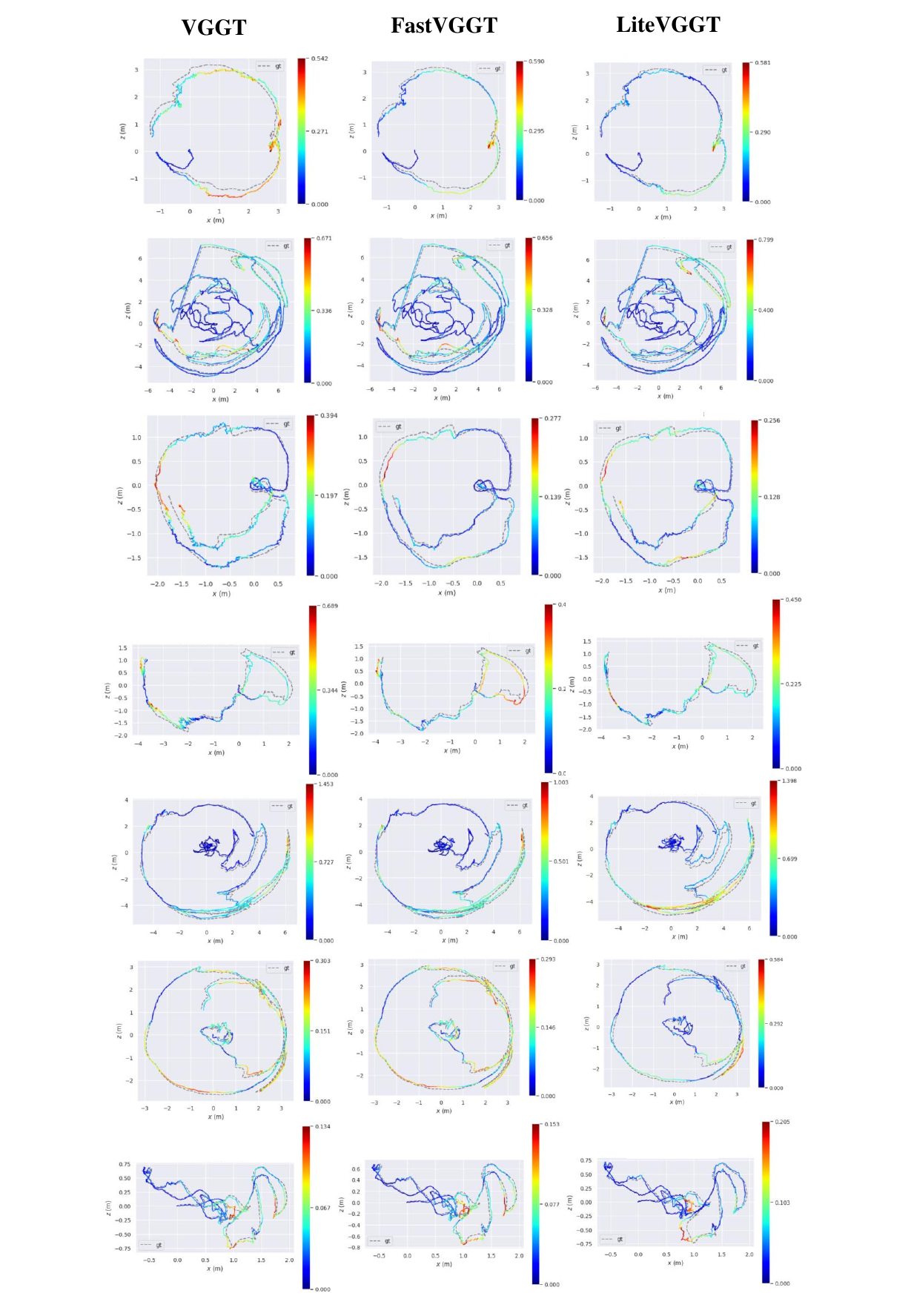}
\vspace{-5mm}
\caption{Additional visualizations of pose estimation results on the ScanNet dataset.}

\label{fig:pose}
\end{figure*}

\begin{figure*}[t]
\centering
\includegraphics[width=1\linewidth]{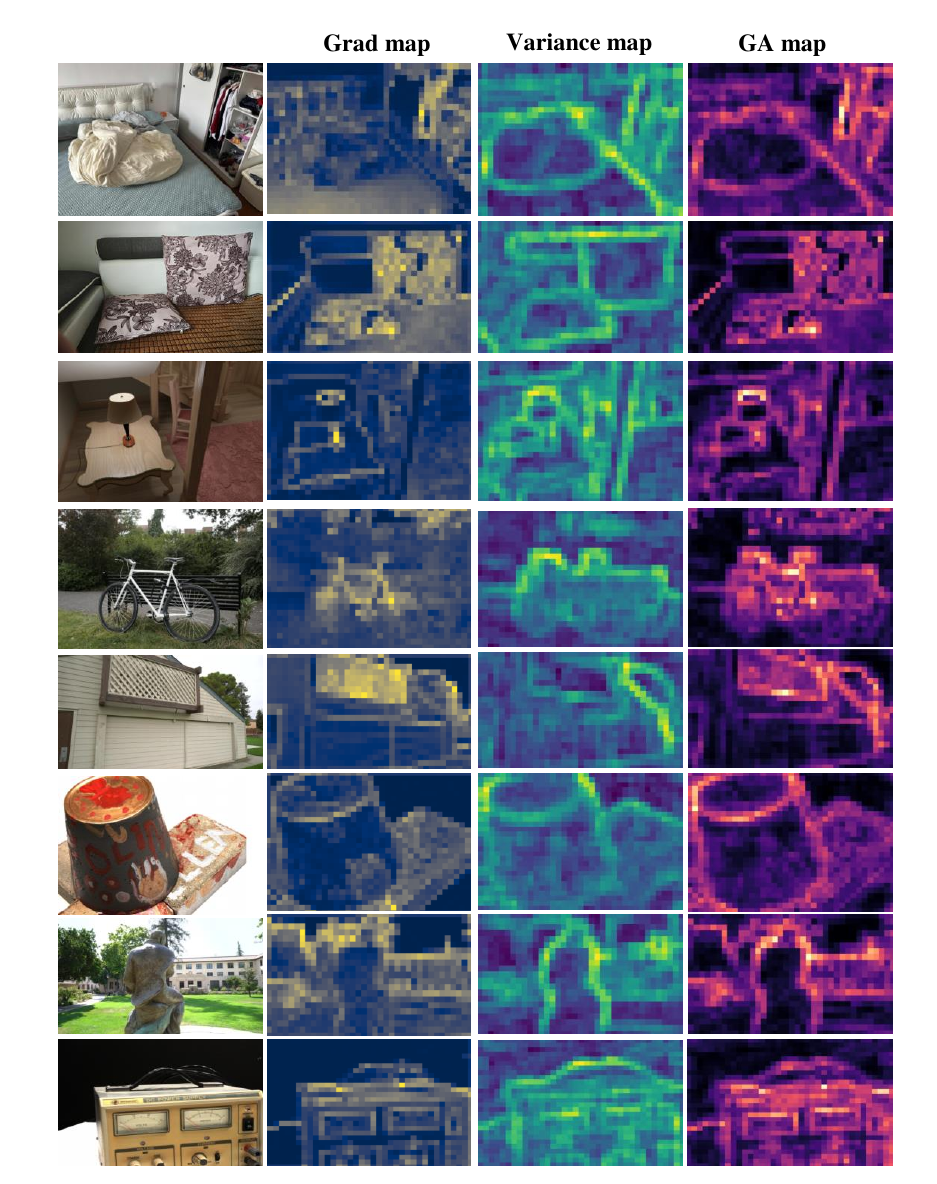}
\vspace{-5mm}
\caption{Additional visualizations of pixel gradients (Grad map), token variance (Variance map), and the fused Geometry-aware map (GA map).}
\label{fig:GA}
\end{figure*}

In Fig. \ref{fig:scene}, we present additional reconstruction results of LiteVGGT, and compare them against the ground-truth (GT) point clouds as well as those produced by VGGT and FastVGGT.
In Fig. \ref{fig:pose}, we show further examples of camera pose estimation, again comparing our results with GT, VGGT, and FastVGGT.
Finally, Fig. \ref{fig:GA} provides additional visualizations of pixel gradients (Grad map), token variance (Variance map), and the fused Geometry-aware map (GA map).

\end{document}